\documentclass[hidelinks,10pt,conference,letterpaper]{IEEEtran}

\IEEEoverridecommandlockouts

\let\labelindent\relax
\usepackage[table,xcdraw]{xcolor}
\usepackage{subfig}
\usepackage{xspace}
\usepackage{graphicx}
\usepackage{ntheorem}
\usepackage{times,amsmath,epsfig,amssymb,graphicx,amsfonts}
\usepackage{empheq}
\usepackage{subfig}
\usepackage{graphicx}
\usepackage{psfrag}
\usepackage{fge}
\usepackage{amsmath}
\usepackage{setspace}
\usepackage{color}
\usepackage{amsfonts}   
\usepackage{amssymb}    
\usepackage{mathrsfs}
\usepackage{setspace}
\usepackage{dblfloatfix}
\usepackage{tikz}
\usepackage{tkz-tab}

\usepackage{pgfplotstable}
\usepackage{pgfplots}
\usepackage{algorithm}
\usepackage{algcompatible}
\usepackage{lipsum}
\usepackage{color}
\usepackage{mathrsfs}
\usepackage{epstopdf}
\usepackage{enumerate}
\usepackage{enumitem}
\usepackage{url}
\usepackage{cite}
\usepackage{tabularx}
\usepackage[pass]{geometry}
\usepackage{chngcntr}

\newcommand{\bdmath}{\begin{dmath}}
\newcommand{\edmath}{\end{dmath}}
\newcommand{\beq}{\begin{equation}}
\newcommand{\eeq}{\end{equation}}
\newcommand{\bdm}{\begin{displaymath}}
\newcommand{\edm}{\end{displaymath}}
\newcommand{\bea}{\begin{eqnarray}}
\newcommand{\eea}{\end{eqnarray}}
\newcommand{\beal}{\beq \begin{array}{lll}}
\newcommand{\eeal}{\end{array} \eeq}
\newcommand{\beas}{\begin{eqnarray*}}
\newcommand{\eeas}{\end{eqnarray*}}
\newcommand{\ba}{\begin{array}}
\newcommand{\ea}{\end{array}}
\newcommand{\bit}{\begin{itemize}}
\newcommand{\eit}{\end{itemize}}
\newcommand{\ben}{\begin{enumerate}}
\newcommand{\een}{\end{enumerate}}

\newcommand{\calA}{{\cal A}}
\newcommand{\calB}{{\cal B}}

\newcommand{\calK}{{\cal K}}

\newcommand{\calM}{{\cal M}}

\newcommand{\calO}{{\cal O}}
\newcommand{\calP}{{\cal P}}

\newcommand{\calR}{{\cal R}}
\newcommand{\calS}{{\cal S}}

\newcommand{\calV}{{\cal V}}

\newcommand{\hide}[1]{}

\newcommand{\hiddenText}{{\color{gray} hidden text.}}
\newcommand{\hideWithText}[1]{\hiddenText}

\newcommand{\diag}[1]{\mathrm{diag}\left(#1\right)}

\newcommand{\eye}{{\mathbf I}}

\newcommand{\Real}[1]{ { {\mathbb R}^{#1} } }

\newcommand{\LQG}{LQG\xspace}
\newcommand{\myParagraph}[1]{{\bf #1.}\xspace}

\newcommand{\tlogdet}{{\tt greedy}\xspace}
\newcommand{\tslqg}{{\tt resilience}\xspace}
\newcommand{\toptimal}{{\tt optimal}\xspace}

\newcommand{\validated}[2]{{#2}}

\newcommand{\elem}{{{v}}}

\floatname{algorithm}{Algorithm}

\floatname{algorithm}{Algorithm}

\newtheorem{mydef}{Definition}

\newtheorem{mytheorem}{Theorem}
\newtheorem{mylemma}{Lemma}
\newtheorem{myremark}{Remark}
\newtheorem{mycorollary}{Corollary}

\newtheorem{myproblem}{Problem}

\newcounter{ale}

\newenvironment{liste}{\begin{itemize}}{\end{itemize}}
\newcommand{\aliste}{\begin{liste} \setcounter{ale}{1}}
\newcommand{\zliste}{\end{liste}}

\title{\huge \bf Resilient Monotone Sequential Maximization}
\author{Vasileios Tzoumas,{$^{1}$} Ali Jadbabaie,{$^{2}$} George J.~Pappas{$^{1}$}
\thanks{$^{1}$The authors are with the Department of Electrical and Systems Engineering, University of Pennsylvania, Philadelphia, PA 19104-6228 USA (email: {\fontsize{8}{8}\selectfont\ttfamily\upshape \{vtzoumas, pappasg\}@seas.upenn.edu}).}
\thanks{$^{2}$The author is with the Institute for Data, Systems and Society, Massachusetts Institute of Technology, Cambridge, MA 02139 USA (email: {\fontsize{8}{8}\selectfont\ttfamily\upshape  jadbabai@mit.edu}).}
\thanks{This work was partially supported  
by the AFOSR Complex Networks program, by the ARL CRA DCIST W911NF-17-2-0181 program, and the Rockefeller Foundation.}
}

\begin{document}
\maketitle

\begin{abstract}
Applications in machine learning, optimization, and control require the sequential selection of a few system elements, such as sensors, data, or actuators, to optimize the system performance across multiple time steps. However, in failure-prone and adversarial environments, sensors get attacked, data get deleted, and actuators fail.  Thence, traditional sequential  design paradigms become insufficient and, in~contrast, \textit{resilient} sequential designs that {adapt} against {system-wide} attacks, deletions, or failures become important. In~general, resilient sequential design problems are computationally hard.  Also, even though they often involve objective functions that are monotone and (possibly) submodular, no scalable approximation algorithms are known for their solution. In~this~paper, we provide the first scalable algorithm, that achieves the following characteristics:~\textit{system-wide resiliency}, i.e., the algorithm is valid for any number of denial-of-service attacks, deletions, or failures;~\textit{adaptiveness}, i.e., at each time step, the algorithm selects system elements based on the history of inflicted attacks, deletions, or failures;~and~\textit{provable approximation performance}, i.e., the algorithm guarantees for monotone objective functions a solution close to the optimal. We quantify the algorithm's approximation performance using a notion of curvature for monotone {(not necessarily submodular) set functions.}
Finally, we~support our theoretical analyses with simulated experiments, by considering a control-aware sensor scheduling scenario, namely, {\textit{sensing-constrained robot navigation}.}
\end{abstract}

\begin{tikzpicture}[overlay, remember picture]
	\path (current page.north east) ++(-3.3,-0.2) node[below left] {
		This paper has been accepted for publication in the IEEE Transactions on Automatic Control.
	};
\end{tikzpicture}
\begin{tikzpicture}[overlay, remember picture]
	\path (current page.north east) ++(-4.8,-0.6) node[below left] {
		Please cite the paper as:
		V.~Tzoumas, A.~Jadbabaie, George J.~Pappas, 
	};
\end{tikzpicture}
\begin{tikzpicture}[overlay, remember picture]
	\path (current page.north east) ++(-2.1,-1) node[below left] {
		``Robust and Adaptive Sequential Submodular Optimization", IEEE Transactions on Automatic Control (TAC), in press.
	};
\end{tikzpicture}

\section{Introduction}\label{sec:Intro}

Problems in machine learning, optimization, and control
\cite{golovin2011adaptive,tzoumas2018codesign,clark2014supermodular,das2011spectral,khanna2017scalable,candes2006stable,boutsidis2009improved,zhao2016scheduling,tokekar2014multi}
require the design of {systems in applications such as:}
\begin{itemize}
\item \textit{Car-congestion prediction}: Given a flood of driving data, collected from the drivers' smart-phones,  which \textit{few} drivers' data should we process 
at \textit{each time of the day}
to enable the accurate prediction of car traffic?~\cite{golovin2011adaptive}
\item \textit{Adversarial-target tracking}: At a flying robot, that uses on-board sensors to navigate itself, which \textit{few} sensors should we activate \textit{at each time step} both to maximize the robot's battery life, and to ensure its ability to track targets moving in a cluttered environment?~\cite{tzoumas2018codesign}
\item \textit{Hazardous environmental-monitoring}: In a team of mobile robots, which \textit{few} robots should we choose \textit{at each time step}  as actuators (leaders)  to guarantee the team's capability to monitor the radiation around a nuclear reactor despite intro-robot communication noise?~\cite{clark2014supermodular}
\end{itemize}
In particular, all the aforementioned applications \cite{clark2014supermodular,golovin2011adaptive,tzoumas2018codesign,das2011spectral,khanna2017scalable,candes2006stable,boutsidis2009improved,zhao2016scheduling,tokekar2014multi} motivate the \textit{sequential} selection of \textit{a few} system elements, such as sensors, data, or actuators, to optimize the system performance across multiple time steps, subject to a resource constraint, such as limited battery for sensor activation.  More formally, each of the above applications motivate the solution to a sequential optimization problem of the form:
\begin{align}\label{eq:non_res}
\begin{split}
&\underset{\mathcal{A}_1 \subseteq \mathcal{V}_1}{\max} \;\cdots
\;\underset{\mathcal{A}_T \subseteq \mathcal{V}_T}{\max}\;\; f(\mathcal{A}_{1},\ldots,\mathcal{A}_T),\\
&\hspace*{1.5cm}\text{such that:}\\
&\hspace*{1.5cm} \text{$|\calA_t|= \alpha_t$, for all $t=1,\ldots,T$\!,}
\end{split}
\end{align}
where $T$ represents the number of design steps in time; the objective function~$f$ is monotone and (possibly) submodular ---submodularity is a diminishing returns property;---
and the cardinality bound $\alpha_t$ captures a resource constraint at time~$t$.
The problem in eq.~\eqref{eq:non_res} is combinatorial, and, specifically, it is NP-hard~\cite{Feige:1998:TLN:285055.285059}; notwithstanding, several approximation algorithms have been proposed \mbox{for its solution, such as the~greedy~\cite{fisher1978analysis}.}

But in all the above critical applications, sensors can get cyber-attacked~\cite{wood2002dos}; data can get deleted~\cite{mirzasoleiman2017deletion}; and actuators can fail~\cite{willsky1976survey}.  Hence, in such failure-prone and adversarial scenarios, 
\textit{resilient} sequential designs that \textit{adapt} 
against
denial-of-service attacks, deletions, or failures
become important. 

In this paper, we formalize for the first time a problem  of \textit{resilient monotone sequential maximization}, that goes beyond the traditional objective of the problem in eq.~\eqref{eq:non_res}, and guards {adaptively} against real-time  attacks, deletions, or failures.
In~particular, we introduce the following resilient re-formulation of the problem in eq.~\eqref{eq:non_res}:
\begin{align}\label{eq:res}
\begin{split}
&\underset{\mathcal{A}_1 \subseteq \mathcal{V}_1}{\max} \;  \underset{\;\mathcal{B}_1 \subseteq \mathcal{A}_1}{\min} \; \cdots
\;\underset{\mathcal{A}_T \subseteq \mathcal{V}_T}{\max} \;  \underset{\mathcal{B}_T \subseteq \mathcal{A}_T}{\min} \; f(\mathcal{A}_{1}\setminus \mathcal{B}_1,\ldots,\mathcal{A}_T\setminus \mathcal{B}_T),\\
&\hspace*{1.5cm}\text{such that:}\\
&\hspace*{1.5cm} \text{$|\calA_t|= \alpha_t$ and $|\calB_t|\leq \beta_t$, for all $t=1,\ldots,T$\!,}
\end{split}
\end{align}
where the number $\beta_t$ represents the number of possible attacks, deletions, or failures ---in general, it is $\beta_t\leq \alpha_t$. Overall, the problem in eq.~\eqref{eq:res}
maximizes the function~$f$ despite real-time \textit{worst-case} failures that compromise the consecutive maximization steps in eq.~\eqref{eq:non_res}. Therefore, the problem formulation in eq.~\eqref{eq:res} is suitable in scenarios where there is no prior on the removal mechanism, as well as, in scenarios where protection against worst-case failures is essential, such as in expensive experiment designs, or missions of adversarial-target tracking.

In more detail, the problem in eq.~\eqref{eq:res} may be interpreted as a $T$-stage perfect information sequential game between two players~\cite[Chapter~4]{myerson2013game}, namely, a ``maximization'' player (designer), and a ``minimization'' player (attacker), who play sequentially, \textit{both observing all past actions of all players}, and with the designer starting the game.  That is, 
at each time $t=1,\ldots,T$\!, both the designer and the attacker \textit{adapt} their set selections to the history of all the players' selections so far, and, in particular, the attacker adapts its selection also to the current ($t$-th) selection of the designer (since at each step $t$, the attacker plays after it observes the selection of the `designer). 

In sum, the problem in eq.~\eqref{eq:res} goes beyond traditional (non-resilient) optimization~\cite{golovin2010adaptive,iyer2013curvature,bian2017guarantees,nemhauser1978best,conforti1984curvature} by proposing \textit{resilient} optimization;
beyond single-step resilient optimization~\cite{orlin2015robust,tzoumas2017resilient,bogunovic2018robust} by proposing \textit{multi-step} (sequential) resilient optimization; 
beyond memoryless resilient optimization~\cite{mitrovic2017streaming} by proposing \textit{adaptive} resilient optimization; 
and beyond protection against {non}-adversarial attacks~\cite{mirzasoleiman2017deletion,kazemi2017deletion} by proposing protection against \textit{worst-case} attacks.  Hence, the problem in eq.~\eqref{eq:res} aims to protect the system performance over extended periods of time against real-time denial-of-service attacks or failures, which is vital in critical applications, such as
 {multi-target surveillance with teams of mobile robots~\cite{tokekar2014multi}.}

\myParagraph{Contributions} In this paper, we make the contributions:
\begin{itemize}
\item (\textit{Problem definition}) We formalize the problem of \textit{resilient monotone sequential maximization} against denial-of-service removals, per eq.~\eqref{eq:res}. This is the first work to formalize, address, and motivate this problem. 
\item (\textit{Solution}) We develop the first algorithm for the problem of resilient monotone sequential maximization in eq.~\eqref{eq:res}, and prove it has the following properties:
\begin{itemize}
\item \textit{system-wide resiliency}: the algorithm is valid for any number of removals;
\item \textit{adaptiveness}: the algorithm adapts the solution to each of the maximization steps in eq.~\eqref{eq:res} to the history of realized (inflicted) removals;
\item \textit{minimal running time}: the algorithm terminates with the same running time as state-of-the-art algorithms for (non-resilient) set function optimization, per eq.~\eqref{eq:non_res};
\item \textit{provable approximation performance}: the algorithm ensures for all $T\geq 1$, and for objective functions $f$ that are monotone and (possibly) submodular 
{---as it holds true in all aforementioned applications~\cite{clark2014supermodular,golovin2011adaptive,tzoumas2018codesign,das2011spectral,khanna2017scalable,candes2006stable,boutsidis2009improved,zhao2016scheduling,tokekar2014multi},---} a solution finitely close to the optimal.  

To quantify the algorithm's approximation performance, we used a notion of curvature for monotone (not necessarily submodular) set functions.
\end{itemize}
\item (\textit{Simulations}) We conduct simulations in a variety of sensor scheduling scenarios for autonomous robot navigation, varying the number of sensor failures. Our simulations validate the benefits of our approach.
\end{itemize}

Overall, the proposed algorithm  herein enables the resilient reformulation and solution of all above applications~\cite{clark2014supermodular,golovin2011adaptive,tzoumas2018codesign,das2011spectral,khanna2017scalable,candes2006stable,boutsidis2009improved,zhao2016scheduling,tokekar2014multi} against worst-case attacks, deletions, or failures, over multiple design steps, {and with provable approximation guarantees.} 

\myParagraph{Notation}   
Calligraphic fonts denote sets (e.g., $\calA$).  Given a set $\calA$, then $2^\calA$ denotes the power set of $\calA$; in addition, $|\calA|$ denotes $\calA$'s cardinality; given also a set $\calB$, then $\calA\setminus\calB$ denotes the set of elements in $\calA$ that are not in~$\calB$. Given a ground set $\mathcal{V}$, a set function $f:2^\mathcal{V}\mapsto \mathbb{R}$, and an element $x\in \mathcal{V}$, the $f(x)$ denotes $f(\{x\})$, and the $f(\calA,\calB)$ denotes $f(\calA\cup\calB)$.  

\section{Resilient Monotone Sequential Maximization}\label{sec:problem_statement}

We formally define resilient monotone sequential maximization.  
We start with the basic definition of monotonicity.

\begin{mydef}[Monotonicity]\label{def:mon}
Consider any finite ground set~$\mathcal{V}$. The set function $f:2^\mathcal{V}\mapsto \mathbb{R}$ is non-decreasing if and only if for any sets $\mathcal{A}\subseteq \mathcal{A}'\subseteq\mathcal{V}$, it holds $f(\mathcal{A})\leq f(\mathcal{A}')$.
\end{mydef}

We define next the main problem in this paper.

\begin{myproblem}\label{pr:robust_sub_max} 
\emph{\textbf{(Resilient Monotone Sequential Maximization)}}
Consider the parameters: an integer $T$; finite ground sets $\mathcal{V}_1,\ldots,\calV_T$;
a non-decreasing set function $f:2^{\mathcal{V}_1}\times \cdots\times 2^{\mathcal{V}_T} \mapsto \mathbb{R}$
such that, without loss of generality, it holds $f(\emptyset)=0$ and $f$ is non-negative;
finally, integers $\alpha_t$ and $\beta_t$ such that $0\leq\beta_t \leq \alpha_t \leq |\mathcal{V}_t|$, for all $t=1,2,\ldots,T$\!.

The problem of \emph{resilient monotone sequential maximization} is to 
maximize the objective function~$f$ through a sequence of~$T$ maximization steps, despite compromises to the solutions of each of the maximization steps; in particular, 
at each maximization step $t=1,\ldots,T$ 
a set $\calA_t\subseteq \calV_t$ of cardinality $\alpha_t$ is selected,
and 
is compromised by a worst-case set removal~$\calB_t$ of cardinality $\beta_t$.
Formally:
\vspace*{1mm}
\begin{align}
\begin{split}
&\underset{\mathcal{A}_1 \subseteq \mathcal{V}_1}{\max} \;  \underset{\;\mathcal{B}_1 \subseteq \mathcal{A}_1}{\min} \; \cdots
\;\underset{\mathcal{A}_T \subseteq \mathcal{V}_T}{\max} \;  \underset{\mathcal{B}_T \subseteq \mathcal{A}_T}{\min} \; f(\mathcal{A}_{1}\setminus \mathcal{B}_1,\ldots,\mathcal{A}_T\setminus \mathcal{B}_T),\\
&\hspace*{1.5cm}\text{such that:}\\
&\hspace*{1.5cm} \text{$|\calA_t|= \alpha_t$ and $|\calB_t|\leq \beta_t$, for all $t=1,\ldots,T$\!.}
\end{split}
\end{align}
\end{myproblem}
\medskip

As we mentioned in this paper's Introduction, Problem~\ref{alg:rob_sub_max} may be interpreted as a $T$-stage perfect information sequential game between two players~\cite[Chapter~4]{myerson2013game}, a ``maximization'' player, and a ``minimization'' player, who play sequentially, {both observing all past actions of all players}, and with the ``maximization'' player starting the game.  In the following paragraphs, we describe this game in more detail:
\begin{itemize}
\item \textit{1st round of the game in Problem~\ref{pr:robust_sub_max}}: the ``maximization'' player selects the set $\calA_1$; then, the ``minimization'' player observes $\calA_1$, and selects the set $\calB_1$ against $\calA_1$;
\item \textit{2nd round of the game in Problem~\ref{pr:robust_sub_max}}: the ``maximization'' player, who already knows $\calA_1$, observes $\calB_1$, and selects the set $\calA_2$, given $\calA_1$ and $\calB_1$; then, the ``minimization'' player, who already knows $\calA_1$ and $\calB_1$, observes $\calA_2$, and selects the set $\calB_2$ against $\calA_2$, given $\calA_1$ and $\calB_1$.

\begin{center}
$\vdots$
\end{center}

\item \textit{$T$-th round of the game in Problem~\ref{pr:robust_sub_max}}: the ``maximization'' player, who already knows the history of selections $\calA_1,\ldots, \calA_{T-1}$, as well as, removals $\calB_{1},\ldots, \calB_{T-1}$, selects the set $\calA_T$, given $\calA_1,\ldots, \calA_{T-1}$ and $\calB_{1},\ldots, \calB_{T-1}$; then, the ``minimization'' player, who also already knows the history of selections $\calA_1,\ldots, \calA_{T-1}$, as well as, removals $\calB_{1},\ldots, \calB_{T-1}$, observes $\calA_T$, and selects the set $\calB_T$ against $\calA_T$, given $\calA_1,\ldots, \calA_{T-1}$ and $\calB_{1},\ldots, \calB_{T-1}$.
\end{itemize} 


\section{Adaptive Algorithm for Problem~\ref{pr:robust_sub_max}} \label{sec:algorithm}

We present the first algorithm for Problem~\ref{pr:robust_sub_max}, show it is adaptive, and, finally, describe the intuition behind it.  The pseudo-code of the algorithm is described in Algorithm~\ref{alg:rob_sub_max}. 

\subsection{Intuition behind Algorithm~\ref{alg:rob_sub_max}}\label{subsec:intuition}

The goal of Problem~\ref{pr:robust_sub_max} is to ensure a maximal value for an objective function $f$ through a sequence of~$T$ maximization steps, despite compromises to the solutions of each of the maximization steps.  In particular, at each maximization step $t=1,\ldots,T$\!, Problem~\ref{pr:robust_sub_max} aims to select a set $\calA_t$ towards a maximal value of $f$\!, despite that each~$\calA_t$ is compromised by a worst-case set removal $\calB_t$ from $\calA_t$, resulting to $f$ being finally evaluated at the sequence of sets $\calA_1\setminus \calB_1,\ldots, \calA_T\setminus \calB_T$ instead of the sequence of sets $\calA_1,\ldots,\calA_T$.
In~this~context, Algorithm~\ref{alg:rob_sub_max} aims to fulfil the goal of Problem~\ref{pr:robust_sub_max} by constructing each set $\calA_t$ as the union of the sets $\calS_{t,1}$, and $\calS_{t,2}$ (line~\ref{line:selection} of Algorithm~\ref{alg:rob_sub_max}), whose role we describe in more detail below:
\setcounter{paragraph}{0} 
\paragraph{Set $\calS_{t,1}$ approximates worst-case set removal from~$\calA_{t}$}  Algorithm~\ref{alg:rob_sub_max} aims with the set $\calS_{t,1}$  to capture the worst-case removal of $\beta_t$ elements among the $\alpha_t$ elements that Algorithm~\ref{alg:rob_sub_max} is going to select in~$\calA_t$; equivalently, the set~$\calS_{t,1}$ is aimed to act as a ``bait'' to an attacker that selects to remove the \textit{best}~$\beta_t$ elements from~$\calA_{t}$ (\textit{best} with respect to the elements' contribution towards the goal of Problem~\ref{pr:robust_sub_max}). However, the problem of selecting the \textit{best}~$\beta_t$ elements in~$\calV_t$ is a combinatorial and, in general,  intractable problem~\cite{Feige:1998:TLN:285055.285059}. 
For this reason, Algorithm~\ref{alg:rob_sub_max} aims to \textit{approximate} the best~$\beta_t$ elements in $\calV_t$, by letting $\calS_{t,1}$ be the set of~$\beta_t$ elements with the largest marginal contributions in the value of the objective function $f$ (lines~\ref{line:sort}-\ref{line:bait} of Algorithm~\ref{alg:rob_sub_max}). 

\paragraph{Set $\calS_{t,2}$ is such that $\calS_{t,1}\cup \calS_{t,2}$ approximates optimal set solution to $t$-th maximization step of Problem~\ref{pr:robust_sub_max}}
Assuming that $\calS_{t,1}$ is the set of $\beta_t$ elements that are going to be removed from Algorithm~\ref{alg:rob_sub_max}'s set selection~$\calA_{t}$,
Algorithm~\ref{alg:rob_sub_max} needs next to select a set $\calS_{t,2}$ of $\alpha_t-\beta_t$ elements to complete the construction of~$\calA_t$, since it is $|\calA_{t}|=\alpha_t$ per Problem~\ref{pr:robust_sub_max}. In~particular, for $\calA_{t}=\calS_{t,1}\cup \calS_{t,2}$ to be an optimal solution to $t$-th maximization step of Problem~\ref{pr:robust_sub_max} (assuming the removal of~$\calS_{t,1}$ from $\calA_t$), Algorithm~\ref{alg:rob_sub_max} needs to select $\calS_{t,2}$ as a \textit{best} set of $\alpha_t-\beta_t$ elements from $\calV_t\setminus\calS_{t,1}$.  
Nevertheless, the problem of selecting a \textit{best} set 
of $\alpha_t-\beta_t$ elements from $\calV_t\setminus\calS_{t,1}$ 
is a combinatorial and, in~general, intractable problem~\cite{Feige:1998:TLN:285055.285059}.  As~a~result, Algorithm~\ref{alg:rob_sub_max} aims to \textit{approximate} such a best set, 
using the greedy procedure in  the lines~\ref{line:begin_while}-\ref{line:end_while} of   Algorithm~\ref{alg:rob_sub_max}.  

Overall, Algorithm~\ref{alg:rob_sub_max} constructs the sets $\calS_{t,1}$ and $\calS_{t,2}$ to approximate an optimal solution $\calA_t$ to the $t$-th maximization step of   Problem~\ref{pr:robust_sub_max} with their union (line~\ref{line:selection} of Algorithm~\ref{alg:rob_sub_max}).

We describe next the steps in Algorithm~\ref{alg:rob_sub_max} in more detail.

\begin{algorithm}[t]
\caption{Adaptive algorithm for Problem~\ref{pr:robust_sub_max}.}
\begin{algorithmic}[1]
\REQUIRE Per Problem~\ref{pr:robust_sub_max}, Algorithm~\ref{alg:rob_sub_max} receives two input types:
\begin{itemize}
\item (\textit{Off-line})~Integer $T$; finite ground sets $\mathcal{V}_1,\ldots,\calV_T$; set function $f:2^{\mathcal{V}_1}\times \cdots\times 2^{\mathcal{V}_T} \mapsto \mathbb{R}$ such that $f$ is non-decreasing, non-negative, and $f(\emptyset)=0$; integers $\alpha_t$ and $\beta_t$ such that $0\leq\beta_t \leq \alpha_t \leq |\mathcal{V}_t|$, for all $t=1,\ldots,T$\!.
\item (\textit{On-line})~At each step $t=2,3,\ldots,T$: realized set removal $\calB_{t-1}$ from Algorithm~\ref{alg:rob_sub_max}'s set selection $\mathcal{A}_{t-1}$.
\end{itemize}
\ENSURE  At each step $t=1,2,\ldots,T$\!, set $\mathcal{A}_{t}$.
\medskip

\FORALL {$t=1,\ldots,T$}\label{line:begin_for_1}
\STATE $\mathcal{S}_{t,1}\leftarrow\emptyset$;~~~$\mathcal{S}_{t,2}\leftarrow\emptyset$;\label{line:initiliaze}
\STATE Sort elements in $\mathcal{V}_t$ such that $\mathcal{V}_t\equiv\{v_{t,1}, \ldots, v_{t,|\calV_t|}\}$
and $f(v_{t,1})\geq \ldots \geq f(v_{t,|\calV_t|})$;\label{line:sort}
\STATE $\mathcal{S}_{t,1}\leftarrow\{v_{t,1}, \ldots, v_{t,\beta}\}$; \label{line:bait}
\WHILE {$|\mathcal{S}_{t,2}| < \alpha_t-\beta_t$} \label{line:begin_while} 
\STATE $x\in \arg\max_{y \in \mathcal{V}_t\setminus (\mathcal{S}_{t,1}\cup\mathcal{S}_{t,2})}f(\calA_{1}\setminus\calB_{1},\ldots,\calA_{t-1}\setminus\calB_{t-1}, \mathcal{S}_{t,2}\cup \{y\})$; \label{line:greedy_1}
\STATE $\mathcal{S}_{t,2} \leftarrow \{x\}\cup \mathcal{S}_{t,2}$;\label{line:greedy_2}
\ENDWHILE \label{line:end_while}
\STATE $\mathcal{A}_{t}\leftarrow \mathcal{S}_{t,1} \cup \mathcal{S}_{t,2}$; \label{line:selection}
\ENDFOR \label{line:end_for_3}

\end{algorithmic}\label{alg:rob_sub_max}
\end{algorithm}

\subsection{Description of steps in Algorithm~\ref{alg:rob_sub_max}}
 
 Algorithm~\ref{alg:rob_sub_max}
 executes four steps for each $t=1,\ldots,T$, where $T$ is the number of maximization steps in Problem~\ref{pr:robust_sub_max}: 
\setcounter{paragraph}{0}
\paragraph{Initialization (line~\ref{line:initiliaze} of Algorithm~\ref{alg:rob_sub_max})} Algorithm~\ref{alg:rob_sub_max} defines two auxiliary sets, namely, the $\calS_{t,1}$ and $\mathcal{S}_{t,2}$, and initializes each of them with the empty set (line~\ref{line:initiliaze} of Algorithm~\ref{alg:rob_sub_max}). The~purpose of $\calS_{t,1}$ and of $\mathcal{S}_{t,2}$ is to construct the set $\calA_{t}$, which is the set Algorithm~\ref{alg:rob_sub_max} selects as a solution to Problem~\ref{pr:robust_sub_max}'s~$t$-th maximization step; in particular, 
the union of $\calS_{t,1}$ and of~$\mathcal{S}_{t,2}$ constructs $\calA_t$ at the end of the $t$-th execution of the algorithm's ``for loop'' (lines~\ref{line:begin_for_1}-\ref{line:end_for_3} of Algorithm~\ref{alg:rob_sub_max}).

\paragraph{Construction of set $\calS_{t,1}$ (lines~\ref{line:sort}-\ref{line:bait} of Algorithm~\ref{alg:rob_sub_max})} Algorithm~\ref{alg:rob_sub_max} constructs the set $\calS_{t,1}$ such that $\calS_{t,1}$ contains~$\beta_t$ elements from the ground set $\calV_t$ and, for any element $s \in \calS_{t,1}$ and any element $s' \notin \calS_{t,1}$, the marginal value of $f(s)$ is at least that of~$f(s')$; that is, among all elements in~$\calV_t$, the set~$\calS_{t,1}$ contains a collection of $\beta_t$ elements that correspond to the highest marginal values of $f$\!.  In detail, Algorithm~\ref{alg:rob_sub_max} constructs~$\calS_{t,1}$ by first sorting and indexing all elements in~$\calV_t$ such that $\mathcal{V}_t=\{v_{t,1}, \ldots, v_{t,|\calV_t|}\}$
and $f(v_{t,1})\geq \ldots \geq f(v_{t,|\calV_t|})$ (line~\ref{line:sort} of Algorithm~\ref{alg:rob_sub_max}), and, then, by including in $\calS_{t,1}$ the fist $\beta_t$ elements among the $\{v_{t,1}, \ldots, v_{t,|\calV_t|}\}$ (line~\ref{line:bait} of Algorithm~\ref{alg:rob_sub_max}).

\paragraph{Construction of set $\calS_{t,2}$ (lines~\ref{line:begin_while}-\ref{line:end_while} of Algorithm~\ref{alg:rob_sub_max})} Algorithm~\ref{alg:rob_sub_max} constructs the set $\calS_{t,2}$ by picking greedily $\alpha_t-\beta_t$ elements from the set $\calV_t\setminus \calS_{t,1}$, and
by accounting for the effect that the history of set selections and removals ($\calA_{1}\setminus\calB_1,\ldots, \calA_{t-1}\setminus\calB_{t-1}$) has on the objective function $f$ of Problem~\ref{pr:robust_sub_max}.
Specifically, the greedy procedure in Algorithm~\ref{alg:rob_sub_max}'s ``while loop''  (lines~\ref{line:begin_while}-\ref{line:end_while} of Algorithm~\ref{alg:rob_sub_max}) selects an element $y\in\mathcal{V}_t\setminus (\mathcal{S}_{t,1}\cup\mathcal{S}_{t,2})$ to add in $\calS_{t,2}$ only if $y$ maximizes the value of  $f(\calA_{1}\setminus\calB_1,\ldots, \calA_{t-1}\setminus\calB_{t-1}, \mathcal{S}_{t,2}\cup \{y\})$.

\paragraph{Construction of set $\calA_{t}$ (line~\ref{line:selection} of Algorithm~\ref{alg:rob_sub_max})}
Algorithm~\ref{alg:rob_sub_max} proposes the set $\calA_t$ as a solution to Problem~\ref{pr:robust_sub_max}'s~$t$-th maximization step.  To this end, Algorithm~\ref{alg:rob_sub_max} constructs $\calA_t$ as the union of the previously constructed sets $\calS_{t,1}$ and~$\mathcal{S}_{t,2}$.  

In sum, Algorithm~\ref{alg:rob_sub_max} enables an adaptive solution of Problem~\ref{pr:robust_sub_max}: for each $t=1,2,\ldots$, Algorithm~\ref{alg:rob_sub_max} constructs a solution set $\calA_t$ to the $t$-th maximization step of Problem~\ref{pr:robust_sub_max} based on both the history of selected solutions up to step $t-1$, namely, the sets $\calA_1,\ldots, \calA_{t-1}$, 
and the corresponding history of set removals from $\calA_1,\ldots, \calA_{t-1}$, namely, the
 $\calB_1,\ldots, \calB_{t-1}$.

\section{Performance Guarantees for Algorithm~\ref{alg:rob_sub_max}}\label{sec:performance}

We quantify Algorithm~\ref{alg:rob_sub_max}'s performance, by bounding its running time, and its approximation performance. To this end, we use the following two notions of curvature for set functions.

\subsection{Curvature and total curvature of non-decreasing functions}\label{sec:total_curvature}
 
We present the notions of \emph{curvature} and of \emph{total curvature} for non-decreasing set functions.  We start by describing the notions of \textit{modularity} and 
 \textit{submodularity} for set functions.

\begin{mydef}[Modularity]\label{def:modular}
Consider any finite set~$\mathcal{V}$.  The set function $g:2^\mathcal{V}\mapsto \mathbb{R}$ is modular if and only if for any set $\mathcal{A}\subseteq \mathcal{V}$, it holds $g(\mathcal{A})=\sum_{\elem\in \mathcal{A}}g(\elem)$.
\end{mydef}

In words, a set function $g:2^\mathcal{V}\mapsto \mathbb{R}$ is modular if through~$g$ all elements in $\mathcal{V}$ cannot substitute each other; in particular, Definition~\ref{def:modular} of modularity implies that for any set $\mathcal{A}\subseteq\mathcal{V}$, and for any element $\elem\in \mathcal{V}\setminus\mathcal{A}$, it holds $g(\{\elem\}\cup\mathcal{A})-g(\mathcal{A})= g(\elem)$.

\begin{mydef}[Submodularity~{\cite[Proposition 2.1]{nemhauser78analysis}}]\label{def:sub}
Consider any finite set $\calV$.  The set function $g:2^\calV\mapsto \mathbb{R}$ is \emph{submodular} if and only if
for any sets $\mathcal{A}\subseteq \validated{\mathcal{A}'}{\mathcal{B}}\subseteq\calV$, and any element $\elem\in \calV$, it \validated{is}{holds}  
$g(\mathcal{A}\cup \{\elem\})\!-\!g(\mathcal{A})\geq g(\validated{\mathcal{A}'}{\mathcal{B}}\cup \{\elem\})\!-\!g(\validated{\mathcal{A}'}{\mathcal{B}})$.
\end{mydef}

Definition~\ref{def:sub} implies that a set function $g:2^\calV\mapsto \mathbb{R}$ is submodular if and only if it satisfies a diminishing returns property where
for any set $\mathcal{A}\subseteq \mathcal{V}$, and for any element $\elem\in \mathcal{V}$, the marginal gain $g(\mathcal{A}\cup \{\elem\})-g(\mathcal{A})$ is~non-increasing. 
In contrast to modularity, submodularity implies that the elements in $\mathcal{V}$ \emph{can} substitute each other, since Definition~\ref{def:sub} of submodularity implies the inequality $g(\{\elem\}\cup\mathcal{A})-g(\mathcal{A})\leq g(\elem)$; that is, in the presence of the set $\mathcal{A}$, the element $\elem$ may lose part of its contribution to the  value of $g(\{x\}\cup\mathcal{A})$.

\begin{mydef}\label{def:curvature}
\emph{\textbf{(Curvature of monotone submodular functions~\cite{conforti1984curvature})}}
Consider a finite set $\mathcal{V}$, and a non-decreasing submodular set function $g:2^\mathcal{V}\mapsto\mathbb{R}$ such that (without loss of generality) for any element $\elem \in \mathcal{V}$, it is  $g(\elem)\neq 0$.  The curvature of $g$ is defined as follows: \begin{equation}\label{eq:curvature}
\kappa_g\triangleq 1-\min_{\elem\in\mathcal{V}}\frac{g(\mathcal{V})-g(\mathcal{V}\setminus\{\elem\})}{g(\elem)}.
\end{equation}
\end{mydef}

Definition~\ref{def:curvature} of curvature implies that for any non-decreasing submodular set function $g:2^\mathcal{V}\mapsto\mathbb{R}$, it holds $0 \leq \kappa_g \leq 1$.  In particular, the value of $\kappa_g$ measures how far~$g$ is from modularity, as we explain next: if $\kappa_g=0$, then for all elements $v\in\mathcal{V}$, it holds $g(\mathcal{V})-g(\mathcal{V}\setminus\{v\})=g(v)$, that is, $g$ is modular. In~contrast, if $\kappa_g=1$, then there exist an element $v\in\mathcal{V}$ such that $g(\mathcal{V})=g(\mathcal{V}\setminus\{v\})$, that is, in the presence of $\mathcal{V}\setminus\{v\}$, $v$~loses all its contribution to the value of $g(\mathcal{V})$.

\begin{mydef}\label{def:total_curvature}
\emph{\textbf{(Total curvature of non-decreasing functions~\cite[Section~8]{sviridenko2017optimal})}}
Consider a finite set $\mathcal{V}$, and a monotone set function $g:2^\mathcal{V}\mapsto\mathbb{R}$.  The total curvature of $g$ is defined as follows: 
\begin{equation}\label{eq:total_curvature}
c_g\triangleq 1-\min_{v\in\mathcal{V}}\min_{\mathcal{A}, \mathcal{B}\subseteq \mathcal{V}\setminus \{v\}}\frac{g(\{v\}\cup\mathcal{A})-g(\mathcal{A})}{g(\{v\}\cup\mathcal{B})-g(\mathcal{B})}.
\end{equation}
\end{mydef}

Definition~\ref{def:total_curvature} of total curvature implies that for any non-decreasing set function  $g:2^\mathcal{V}\mapsto\mathbb{R}$, it holds $0 \leq c_g \leq 1$. To connect the notion of total curvature with that of curvature, we note that when the function $g$ is non-decreasing and submodular, then the two notions coincide, i.e., it holds $c_g=\kappa_g$; the reason is that if $g$ is non-decreasing and submodular, then the inner minimum in eq.~\eqref{eq:total_curvature} is attained for $\calA=\calB\setminus\{v\}$ and $\calB=\emptyset$.  
In addition, to connect the above notion of total curvature with the notion of modularity, we note that 
if $c_g=0$, then $g$ is modular, since eq.~\eqref{eq:total_curvature} implies that for any elements $\elem\in\calV$, and for any sets $\calA,\calB\subseteq \calV\setminus \{\elem\}$, it holds: 
\begin{equation}\label{eq:ineq_total_curvature}
(1-c_g)\left[g(\{\elem\}\cup\calB)-g(\calB)\right]\leq g(\{\elem\}\cup\calA)-g(\calA),
\end{equation}
which for $c_g=0$ implies the modularity of $g$. Finally, 
to connect the above notion of total curvature with the notion of monotonicity, we mention that 
if $c_g=1$, then eq.~\eqref{eq:ineq_total_curvature} implies that $g$ is merely non-decreasing  (as it is already assumed by the Definition~\ref{def:total_curvature} of total curvature).

\begin{mydef}[Approximate submodularity]\label{def:approx_sub}
Consider a finite set $\mathcal{V}$, and a non-decreasing set function $g:2^\mathcal{V}\mapsto\mathbb{R}$, whose total curvature $c_g$ is such that $c_g<1$. Then, we say that $g$ is \emph{approximately submodular}.
\end{mydef}

\subsection{Performance analysis for Algorithm~\ref{alg:rob_sub_max}}

We quantify Algorithm~\ref{alg:rob_sub_max}'s approximation performance, as well as, its running time per maximization step in Problem~\ref{pr:robust_sub_max}.

\begin{mytheorem}[Performance of Algorithm~\ref{alg:rob_sub_max}]\label{th:alg_rob_sub_max_performance}
Consider an instance of Problem~\ref{pr:robust_sub_max}, the notation therein, the notation in Algorithm~\ref{alg:rob_sub_max}, and the definitions:
\begin{itemize}
\item let the number $f^\star$ be the (optimal) value to Problem~\ref{pr:robust_sub_max};
\item given sets $\mathcal{A}_{1:T}\triangleq(\calA_1,\ldots, \calA_{T})$ as solutions to the maximization steps in Problem~\ref{pr:robust_sub_max}, let  $\mathcal{B}^\star(\mathcal{A}_{1:T})$ be the collection of optimal (worst-case) set removals from each of the $\mathcal{A}_t$, where $t=1,\ldots,T$\!, per Problem~\ref{pr:robust_sub_max}, i.e.: 
\begin{align*}
&\mathcal{B}^\star(\mathcal{A}_{1:T})\in\arg\min_{\mathcal{B}_1 \subseteq \mathcal{A}_1, |\mathcal{B}_1| \leq \beta_1}\cdots \min_{\mathcal{B}_T \subseteq \mathcal{A}_T, |\mathcal{B}_T| \leq \beta_T}\\
&\hspace{4cm}f(\calA_1\setminus\calB_1,\ldots,\mathcal{A}_T\setminus \mathcal{B}_T);
\end{align*}
\end{itemize}

The performance of Algorithm~\ref{alg:rob_sub_max} is bounded as follows:

\begin{enumerate}[leftmargin=*]
\item \emph{(Approximation performance)}~Algorithm~\ref{alg:rob_sub_max} returns the sequence of sets $\calA_{1:T}\triangleq (\calA_1,\ldots,\calA_T)$ such that, for all $t=1,\ldots,T$\!, it holds $\mathcal{A}_t\subseteq \calV_t$, $|\mathcal{A}_t|\leq \alpha_t$, and:
\begin{itemize}
\item if the objective function $f$ is non-decreasing and submodular, then: 
\begin{equation}\label{ineq:bound_sub}
\frac{f(\mathcal{A}_{1:T}\setminus \mathcal{B}^\star(\calA_{1:T}))}{f^\star}\geq  (1-\kappa_f)^4\!,
 \end{equation}
where $\kappa_f$ is the curvature of $f$ (Definition~\ref{def:curvature}).
    
\item if the objective function $f$ is non-decreasing, then:
\begin{equation}\label{ineq:bound_non_sub}
\frac{f(\mathcal{A}_{1:T}\setminus \mathcal{B}^\star(\calA_{1:T}))}{f^\star}\geq  (1-c_f)^5\!,
\end{equation}
where $c_f$ is the total curvature of $f$ (Definition~\ref{def:total_curvature}).
\end{itemize}

\item \emph{(Running time)}~Algorithm~\ref{alg:rob_sub_max} constructs each set $\calA_t$, for each $t=1,\ldots,T$\!, to solve the $t$-th maximization step of Problem~\ref{pr:robust_sub_max}, with $O(|\mathcal{V}_t|(\alpha_t-\beta_t))$ evaluations of $f$\!. 
\end{enumerate}
\end{mytheorem}


\myParagraph{Provable approximation performance}
Theorem~\ref{th:alg_rob_sub_max_performance}  implies on the approximation performance of Algorithm~\ref{alg:rob_sub_max}:
\setcounter{paragraph}{0}
\paragraph{{Near-optimality}} Both for monotone submodular objective functions $f$  with curvature $\kappa_f<1$, and for merely monotone objective functions~$f$ with total curvature $c_f<1$, Algorithm~\ref{alg:rob_sub_max} guarantees a value for Problem~\ref{pr:robust_sub_max} finitely close to the optimal.  In~particular,  per ineq.~\eqref{ineq:bound_sub} (case of submodular objective functions), the~approximation factor of Algorithm~\ref{alg:rob_sub_max} is bounded by $(1-\kappa_f)^4$\!, which is non-zero for any monotone submodular function~$f$ with $\kappa_f<1$;
similarly, per ineq.~\eqref{ineq:bound_non_sub} (case of approximately-submodular functions), the approximation factor of Algorithm~\ref{alg:rob_sub_max} is bounded by $(1-c_f)^5$\!, which is non-zero for any monotone function~$f$ with $c_f<1$ ---notably, although it is known for the problem of (non-resilient) set function maximization that the approximation bound $(1-c_f)$ is tight~\cite[Theorem~8.6]{sviridenko2017optimal}, the tightness of the bound $(1-c_f)^5$ in ineq.~\eqref{ineq:bound_non_sub} for  Problem~\ref{pr:robust_sub_max} is an open problem.

We discuss classes of functions $f$ with curvatures $\kappa_f<1$ or $c_f<1$, along with relevant applications, in the remark below.

\begin{myremark}\emph{\textbf{(Classes of functions $f$ with $\kappa_f<1$ or $c_f<1$, and applications)}}
\emph{Classes of functions $f$ with $\kappa_f<1$} are the concave over modular functions~\cite[Section~2.1]{iyer2013curvature}, and the $\log\det$ of positive-definite matrices~\cite{sviridenko2015optimal,sharma2015greedy}. \emph{Classes of functions $f$ with $c_f<1$} are support selection functions~\cite{elenberg2016restricted}, 
and estimation error metrics such as the average minimum square error of the Kalman filter~\cite[Theorem~4]{tzoumas2018codesign}

The aforementioned classes of functions $f$ with $\kappa_f<1$ or $c_f<1$ appear in applications of facility location, machine learning, and control, such as sparse approximation and feature selection~\cite{das2011spectral,khanna2017scalable}, sparse
recovery and column subset selection~\cite{candes2006stable,boutsidis2009improved}, and actuator and sensor scheduling~\cite{zhao2016scheduling,tzoumas2018codesign}; as a result, Problem~\ref{pr:robust_sub_max} enables applications such as resilient experiment design, resilient actuator scheduling for minimal control effort, and resilient multi-robot navigation with minimal sensing and communication.
\end{myremark}

\paragraph{Approximation performance for low curvature}
For both monotone submodular and merely monotone objective functions $f$\!,  when the curvature $\kappa_f$ and the total curvature $c_f$, respectively, tend to zero, Algorithm~\ref{alg:rob_sub_max} becomes exact 
since for $\kappa_f\rightarrow 0$ and $c_f\rightarrow 0$ the terms $(1-\kappa_f)^4$ and $(1-c_f)^5$ in ineq.~\eqref{ineq:bound_sub} and ineq.~\eqref{ineq:bound_non_sub} tend to $1$.
Overall, Algorithm~\ref{alg:rob_sub_max}'s curvature-dependent
approximation bounds make a first step towards separating
the classes of monotone submodular and merely monotone  functions into
functions for which Problem~\ref{pr:robust_sub_max}
can be approximated well (low curvature functions), and functions for which it cannot \mbox{(high curvature functions).}

A machine learning  problem where Algorithm~\ref{alg:rob_sub_max} guarantees an approximation performance close to $100\%$ the optimal is that of Gaussian process regression for processes with RBF kernels~\cite{krause2008near,bishop2006pattern}; this problem emerges in applications of sensor deployment and scheduling for temperature monitoring.  The reason that in this class of regression problems Algorithm~\ref{alg:rob_sub_max} performs almost optimally is that the involved objective function is the entropy of the selected sensor measurements, which for Gaussian processes with RBF kernels has curvature value close to zero~\cite[Theorem~5]{sharma2015greedy}. 

\paragraph{Approximation performance for no failures or attacks}
Both for monotone submodular objective functions~$f$\!, and for merely monotone objective functions $f$\!, when the number of attacks, deletions, and failures is zero ($\beta_t=0$, for all $t=1,\ldots,T$), Algorithm~\ref{alg:rob_sub_max}'s approximation performance is the same as that of the state-of-the-art algorithms for (non-resilient) set function maximization.  In particular, when for all $t=1,\ldots,T$ it is $\beta_t=0$, then Algorithm~\ref{alg:rob_sub_max} is the same as the local greedy algorithm, proposed in~\cite[Section~4]{fisher1978analysis} for (non-resilient) set function maximization, whose approximation performance is optimal~\cite[Theorem~8.6]{sviridenko2017optimal}.

\myParagraph{Minimal running time}
Theorem~\ref{th:alg_rob_sub_max_performance} implies that Algorithm~\ref{alg:rob_sub_max}, even though it goes beyond the objective of (non-resilient) multi-step set function optimization, by accounting for attacks, deletions, and failures, it has the same order of running time as state-of-the-art algorithms for (non-resilient) multi-step set function optimization. In particular, such algorithms for (non-resilient) multi-step set function optimization~\cite[Section~4]{fisher1978analysis}~\cite[Section~8]{sviridenko2017optimal} terminate with $O(|\calV_t|(\alpha_t-\beta_t))$ evaluations of the objective function $f$ per maximization step $t=1,\ldots,T$, and Algorithm~\ref{alg:rob_sub_max} also terminates with $O(|\calV_t|(\alpha_t-\beta_t))$ evaluations of the objective function~$f$ per maximization step  $t=1,\ldots,T$.

\myParagraph{Summary of theoretical results} In sum, Algorithm~\ref{alg:rob_sub_max} is the first algorithm for Problem~\ref{pr:robust_sub_max}, and it enjoys:
\begin{itemize}
\item \textit{system-wide resiliency}: Algorithm~\ref{alg:rob_sub_max} is valid for any number of denial-of-service attacks, deletions, and failures;
\item \textit{adaptiveness}: Algorithm~\ref{alg:rob_sub_max} adapts the solution to each of the maximization steps in Problem~\ref{pr:robust_sub_max} to the history of inflicted denial-of-service attacks and failures;
\item \textit{minimal running time}:  Algorithm~\ref{alg:rob_sub_max} terminates with the same running time as state-of-the-art algorithms for (non-resilient) multi-step submodular function optimization;
\item \textit{provable approximation performance}: Algorithm~\ref{alg:rob_sub_max} ensures for all monotone objective functions $f$ that are either submodular or approximately submodular ($c_f<1$), and for all $T\geq 1$, a solution finitely close to the optimal.
\end{itemize}

Notably, Algorithm~\ref{alg:rob_sub_max} is also the first to guarantee for any number of failures, and for monotone approximately-submodular functions~$f$\!, a provable approximation performance for the {one-step} version {of Problem~\ref{pr:robust_sub_max} where $T=1$.}

\section{Numerical Experiments}\label{sec:simulations}

In this section, we demonstrate a near-optimal performance of Algorithm~\ref{alg:rob_sub_max} via numerical experiments.  In particular, we consider a control-aware sensor scheduling scenario, namely,  \textit{sensing-constrained robot navigation}.\footnote{The scenario of {sensing-constrained robot navigation} is introduced in~\cite[Section~V.B]{tzoumas2018codesign}, in the absence of sensor failures.} 
According to this scenario, an unmanned aerial vehicle (UAV), which has limited remaining battery and measurement-processing power, has the objective to land, and to this end, it schedules to activate at each time step only a subset of its on-board sensors, so to localize itself and enable the generation of a control input for landing; specifically, at each time step, the UAV generates its control input by implementing an LQG-optimal controller, given the measurements collected by the activated sensors up to the current time step~\cite{tzoumas2018codesign,bertsekas2005dynamic}.

In more detail, in the following paragraphs we present a Monte Carlo analysis for an instance of the aforementioned sensing-constrained robot navigation scenario, in the presence of worst-case sensor failures, and observe that Algorithm~\ref{alg:rob_sub_max} results to a near-optimal sensor selection schedule: in particular, the resulting navigation performance of the UAV matches the optimal in all tested instances for which the optimal selection could be computed via a brute-force approach.

%
%

\myParagraph{Simulation setup}  We adopt the same instance of the sensing-constrained robot navigation scenario adopted in~\cite[Section~V.B]{tzoumas2018codesign}.  Specifically,
a UAV moves in a 3D space, starting from a
randomly selected initial location. 
The objective of the UAV is to land at 
 position $[0,\;0,\;0]$ with zero velocity. 
The UAV is modelled as a double-integrator
with state $x_t = [p_t \; v_t]^\top \in \Real{6}$ at each time $t=1,2,\ldots$  
 ($p_t$ is the 3D position of the UAV, and $v_t$ is its velocity), and can control its own acceleration 
$u_t \in \Real{3}$; the process noise is chosen as $W_t = \eye_6$. 
The UAV is equipped with multiple sensors, as follows: it has two on-board GPS receivers, measuring the 
UAV position  $p_t$ with a covariance $2 \cdot\eye_3$, 
and an altimeter, measuring only the last component of $p_t$ (altitude) with standard deviation $0.5\rm{m}$. 
Moreover, the UAV 
can use a stereo camera to measure the relative position of $10$ landmarks on the ground;
we assume the location of each landmark to be known 
only approximately, and we associate to each landmark an uncertainty covariance,
which is randomly generated at the 
beginning of each run.  
The UAV has limited on-board resource-constraints, hence it can only activate a subset of sensors (possibly different at each time step).
For instance, the resource-constraints may be due to the power consumption of the GPS and the altimeter,
or due to computational constraints that prevent to run object-detection algorithms to detect all landmarks on the ground.  

Among the aforementioned $13$ possible sensor measurements available to the UAV at each time step, we assume that the UAV can use only $\alpha=11$ of them. 
In particular, the UAV chooses the sensors to activate at each time step so to minimize an LQG cost with cost matrices $Q$ (which penalizes the state vector) and $R$ (which penalizes the control input vector), per the problem formulation in~\cite[Section~II]{tzoumas2018codesign};
 specifically, in this simulation setup we set $Q = \diag{[1e^{-3},\; 1e^{-3},\;10,\; 1e^{-3},\; 1e^{-3},\; 10]}$ 
and $R = \eye_3$.  Note that the structure of $Q$ (which penalizes the magnitude of the state vector) reflects the fact that during landing 
we are particularly interested in controlling the vertical direction and the vertical velocity 
(entries with larger weight in $Q$), while we are less interested in controlling accurately the 
horizontal position and velocity (assuming a sufficiently large landing site).  Given a time horizon $T$ for landing, in~\cite{tzoumas2018codesign} it is proven that the UAV  selects an optimal sensor schedule and generates an optimal LQG control input with cost matrices $Q$ and $R$ if it selects the sensors set $\calS_t$ to activate at each time $t=1,\ldots,T$ by minimizing an objective function of the form:
\begin{equation}\label{eq:opt_sensors}
\sum_{t=1}^{T}\text{trace}[M_t\Sigma_{t|t}(\calS_1,\ldots,\calS_t)],
\end{equation}
where $M_t$ is a positive semi-definite matrix that depends on the LQG cost matrices $Q$ and $R$, as well as, on the UAV's model dynamics; and $\Sigma_{t|t}(\calS_1,\ldots,\calS_t)$ is the error covariance of the Kalman filter given the sensor selections up to time $t$.

In the remaining paragraphs, we present results averaged over 10 Monte Carlo runs of the above simulation setup, where in each run we~randomize the 
covariances describing the landmark position uncertainty, and where we vary the number~$\beta$ of sensors failures at each time step $t$: in~particular, we consider  $\beta$ to vary among the values $1,4,7,10$.

\myParagraph{Compared algorithms}
We compare four algorithms. All algorithms
only differ in how they select the sensors used.
The~first algorithm is the optimal sensor selection algorithm, denoted as \toptimal, which 
attains the minimum of the cost function in eq.~\eqref{eq:opt_sensors}; this brute-force approach is viable since the number of available sensors is small.
The second approach is a pseudo-random sensor selection, denoted as {\tt random$^*$}\!\!, 
which selects one of the GPS measurements and a random subset of the lidar measurements; 
note that we do not consider a fully random selection since in practice this 
often leads to an unobservable system.
The third approach, denoted as \tlogdet, selects sensors to greedily minimize the cost function in eq.~\eqref{eq:opt_sensors}, \textit{ignoring the possibility of sensor failures}, per the problem formulation in eq.~\eqref{eq:non_res}.
The fourth approach uses Algorithm~\ref{alg:rob_sub_max} to solve the resilient reformulation of eq.~\eqref{eq:opt_sensors}, per Problem~\ref{pr:robust_sub_max}, and is denoted as \tslqg.  

At each time step, from each of the selected sensor sets, selected by any of the above four algorithms, we consider an optimal sensor removal, which we compute via a brute-force.

\myParagraph{Results}  The results of our numerical analysis are reported in Fig.~\ref{fig:formationControlStats}.  In particular, Fig.~\ref{fig:formationControlStats} shows the \LQG cost for increasing time, for the case where the number of selected sensors at each time step is $\alpha=11$, while the number of sensor failures $\beta$ at each time step varies across the values $10,7,4$, $1$.  The following observations are due: 
\begin{itemize}
\item (\textit{Near-optimality of Algorithm~\ref{alg:rob_sub_max}}) Algorithm~\ref{alg:rob_sub_max} (blue colour in Fig.~\ref{fig:formationControlStats}) performs close to  the optimal brute-force algorithm (green colour in Fig.~\ref{fig:formationControlStats}); in particular, across all scenarios in Fig.~\ref{fig:formationControlStats}, Algorithm~\ref{alg:rob_sub_max} achieves an approximation performance at least 97\% the optimal.
\item (\textit{Performance of greedy algorithm}) The greedy algorithm (red colour in Fig.~\ref{fig:formationControlStats}) performs poorly as the number~$\beta$ of sensor failures increases, which was expected, given that this algorithm greedily minimizes the cost function in eq.~\eqref{eq:opt_sensors} {ignoring the possibility of sensor failures}.
\item (\textit{Performance of random algorithm}) Expectedly, also the performance of the random algorithm (black colour in Fig.~\ref{fig:formationControlStats}) is poor across all scenarios in Fig.~\ref{fig:formationControlStats}.
\end{itemize}

\newcommand{\myhspace}{\hspace{-2mm}}
\newcommand{\mpw}{4.5cm}
\begin{figure}[t]
\myhspace\myhspace
\begin{minipage}{\textwidth}
\begin{tabular}{cc}%
\myhspace
\begin{minipage}{\mpw}%
\centering
\includegraphics[width=1.1\columnwidth]{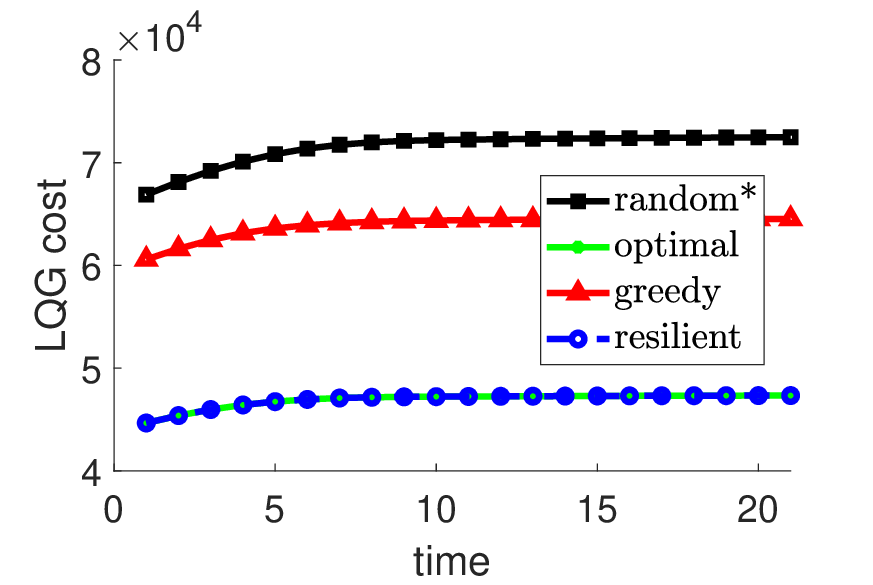} \\
(a) \;\;\;$\beta=10$, $\alpha= 11$
\end{minipage}
& \myhspace
\begin{minipage}{\mpw}%
\centering%
\includegraphics[width=1.1\columnwidth]{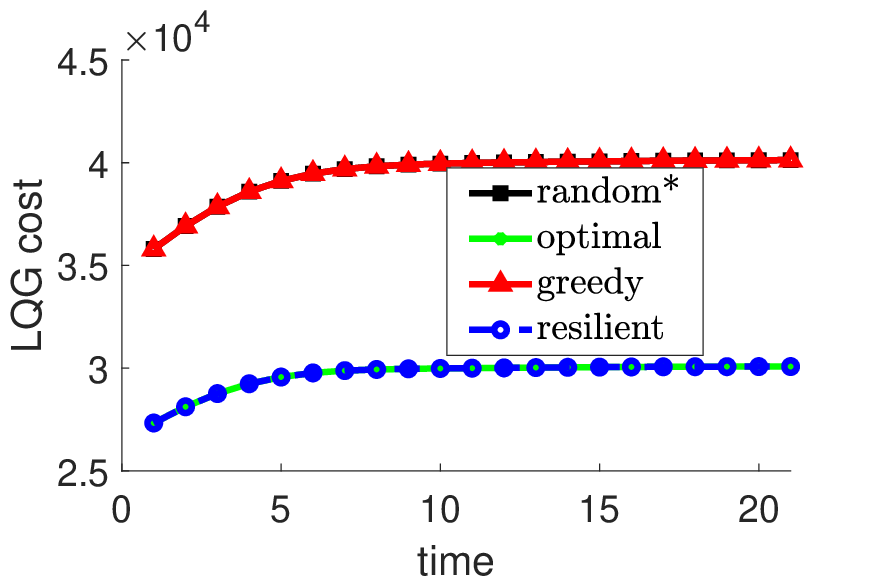} \\
(b) \;\;\;$\beta=7$, $\alpha= 11$  
\end{minipage}
\\
\myhspace
\begin{minipage}{\mpw}%
\centering
\includegraphics[width=1.1\columnwidth]{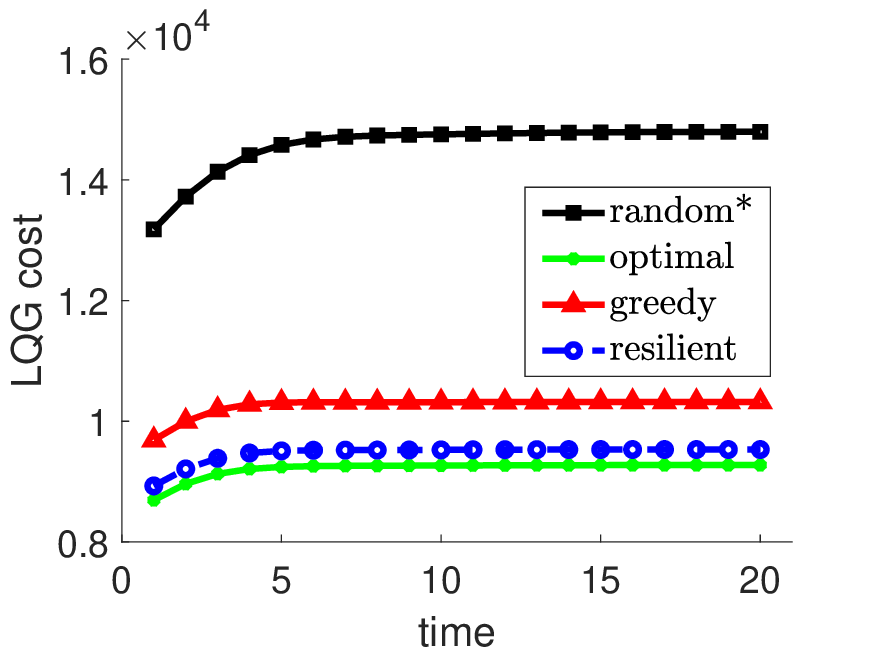} \\
(c)\;\;\;$\beta=4$, $\alpha= 11$ 
\end{minipage}
& \myhspace
\begin{minipage}{\mpw}%
\centering%
\includegraphics[width=1.04\columnwidth]{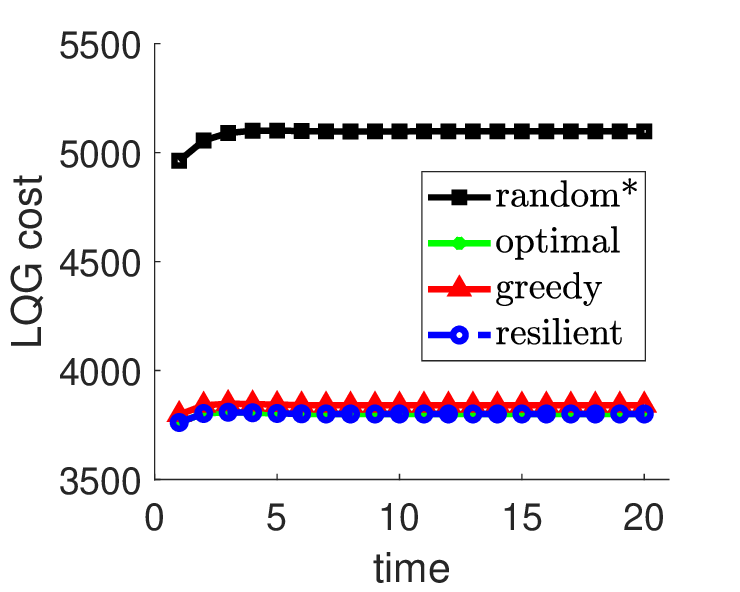} \\
(d)\;\;\; $\beta=1$, $\alpha= 11$ 
\end{minipage}
\end{tabular}
\end{minipage}%
\vspace{-1mm}
\caption{\label{fig:formationControlStats}\small
\LQG cost for increasing time, where across all sub-figures (a)-(d) it is $\alpha=11$ (number of active sensors per time step). The value of $\beta$ (number of sensor failures at each time step among the $\alpha$ active sensors) varies across the sub-figures.
}\vspace{-5mm}
\end{figure}

Overall, in the above numerical experiments, Algorithm~\ref{alg:rob_sub_max} demonstrates a near-optimal approximation performance, and the necessity for the resilient reformulation of the problem in eq.~\eqref{eq:non_res} per Problem~\ref{pr:robust_sub_max} is exemplified.


\section{Concluding remarks \& Future work} \label{sec:con}

We made the first step to ensure the success of critical missions in machine learning and control, that involve the optimization of systems across multiple time-steps, against persistent failures or denial-of-service attacks.  In particular,  we provided the first algorithm for Problem~\ref{pr:robust_sub_max}, which, with minimal running time, adapts to the history of the inflicted failures and attacks, and guarantees
a close-to-optimal performance against system-wide failures and attacks.  To quantify the algorithm's approximation performance, we exploited a notion of curvature for monotone (not necessarily submodular) set functions, and contributed a first step towards characterizing the curvature's effect on the approximability of resilient \textit{sequential} maximization.  Our curvature-dependent characterizations complement the current knowledge on the curvature's effect on the approximability of simpler problems, such as of {non-sequential} resilient maximization~\cite{tzoumas2017resilient,bogunovic2018robust}, and of {non-resilient} maximization~\cite{conforti1984curvature,iyer2013curvature,bian2017guarantees}. Finally, we supported our theoretical analyses with simulated experiments.

This paper opens several avenues for future research, both in theory and in applications.
Future work in theory includes the extension of our results to matroid constraints, to enable applications of set coverage and of network design~\cite{calinescu2011maximizing,clark2015scalable}.  
Future work in applications includes the experimental testing of the proposed algorithm in applications of motion-planning for multi-target covering with mobile vehicles~\cite{tokekar2014multi},
and in applications of control-aware sensor scheduling for multi-agent autonomous navigation~\cite{tzoumas2018codesign}, to enable resiliency in {critical scenarios of surveillance, and of search and rescue.}

\section{Acknowledgements}

We thank Luca Carlone for inspiring discussions, and for sharing code that enabled our numerical experiments.

\bibliographystyle{IEEEtran}
\bibliography{references,security_references}

\appendices

\section{Notation}\label{app:notation}

In the appendix we use the following notation
to support the proofs in this paper: given a finite ground set $\calV$, and a set function $f:2^\mathcal{V}\mapsto \mathbb{R}$, then, for any sets $\mathcal{X}\subseteq \mathcal{V}$ and $\mathcal{X}'\subseteq \mathcal{V}$: 
\begin{equation}\label{notation:marginal}
f(\mathcal{X}|\mathcal{X}')\triangleq f(\mathcal{X}\cup\mathcal{X}')-f(\mathcal{X}').
\end{equation}
Moreover,  let the sets $\mathcal{A}^\star_{1:T}=(\calA^\star_1,\ldots,\calA^\star_T)$ denote an (optimal) solution to Problem~\ref{pr:robust_sub_max}, i.e.:
\begin{align}
\begin{split}
&\mathcal{A}^\star_{1:T}\in \\
&\arg\underset{\mathcal{A}_1 \subseteq \mathcal{V}_1}{\max}  \underset{\;\mathcal{B}_1 \subseteq \mathcal{A}_1}{\min} \cdots
\underset{\mathcal{A}_T \subseteq \mathcal{V}_T}{\max} \underset{\;\mathcal{B}_T \subseteq \mathcal{A}_T}{\min} f(\mathcal{A}_{1}\setminus \mathcal{B}_1,\ldots,\mathcal{A}_T\setminus \mathcal{B}_T),\\
&\hspace*{1.5cm}\text{such that:}\\
&\hspace*{1.5cm} \text{$|\calA_t|= \alpha_t$ and $|\calB_t|\leq \beta_t$, for all $t=1,\ldots,T$\!.}
\end{split}
\end{align}

\section{Preliminary lemmas}\label{app:prelim}

We list lemmas that support the proof of Theorem~\ref{th:alg_rob_sub_max_performance}.

\begin{mylemma}\label{lem:non_total_curvature}
Consider a finite ground set $\mathcal{V}$ and a non-decreasing submodular set function $f:2^\mathcal{V}\mapsto \mathbb{R}$ such that $f$ is non-negative and $f(\emptyset)=0$. Then, for any $\mathcal{A}\subseteq \mathcal{V}$, it~holds:
\begin{equation*}
f(\mathcal{A})\geq (1-\kappa_f)\sum_{a \in \mathcal{A}}f(a).
\end{equation*}
\end{mylemma}
\paragraph*{Proof of Lemma~\ref{lem:non_total_curvature}} Let $\mathcal{A}=\{a_1,a_2,\ldots, a_{|{\cal A}|}\}$. We prove Lemma~\ref{lem:non_total_curvature} by proving the following two inequalities: 
\begin{align}
f(\mathcal{A})&\geq \sum_{i=1}^{|{\cal A}|} f(a_i|\mathcal{V}\setminus\{a_i\}),\label{ineq5:aux_5}\\
\sum_{i=1}^{|{\cal A}|} f(a_i|\mathcal{V}\setminus\{a_i\})&\geq (1-\kappa_f)\sum_{i=1}^{|{\cal A}|} f(a_i)\label{ineq5:aux_6}. 
\end{align} 

We begin with the proof of ineq.~\eqref{ineq5:aux_5}: 
\begin{align}
f(\mathcal{A})&=f(\mathcal{A}|\emptyset)\label{ineq5:aux_9}\\
&\geq f(\mathcal{A}|\mathcal{V}\setminus \mathcal{A})\label{ineq5:aux_10}\\
&= \sum_{i=1}^{|{\cal A}|}f(a_i|\mathcal{V}\setminus\{a_i,a_{i+1},\ldots,a_{|{\cal A}|}\})\label{ineq5:aux_11}\\
&\geq \sum_{i=1}^{|{\cal A}|}f(a_i|\mathcal{V}\setminus\{a_i\}),\label{ineq5:aux_12}
\end{align}
where ineqs.~\eqref{ineq5:aux_10} to~\eqref{ineq5:aux_12} hold for the following reasons: ineq.~\eqref{ineq5:aux_10} is implied by eq.~\eqref{ineq5:aux_9} because $f$ is submodular and $\emptyset\subseteq \mathcal{V}\setminus \mathcal{A}$; eq.~\eqref{ineq5:aux_11} holds since for any sets $\mathcal{X}\subseteq \mathcal{V}$ and $\mathcal{Y}\subseteq \mathcal{V}$ it is $f(\mathcal{X}|\mathcal{Y})=f(\mathcal{X}\cup \mathcal{Y})-f(\mathcal{Y})$, and it also  $\{a_1,a_2,\ldots, a_{|{\cal A}|}\}$ denotes the set $\mathcal{A}$; and ineq.~\eqref{ineq5:aux_12} holds since $f$ is submodular and $\mathcal{V}\setminus\{a_i,a_{i+1},\ldots,a_{\mu}\} \subseteq \mathcal{V}\setminus\{a_i\}$.  These observations complete the proof of ineq.~\eqref{ineq5:aux_5}.

We now prove ineq.~\eqref{ineq5:aux_6} using the Definition~\ref{def:curvature} of $\kappa_f$, as follows: since $\kappa_f=1-\min_{v\in \mathcal{V}}\frac{f(v|\mathcal{V}\setminus\{v\})}{f(v)}$, it is implied that for all elements $v\in \mathcal{V}$ it is $ f(v|\mathcal{V}\setminus\{v\})\geq (1-\kappa_f)f(v)$.  Therefore, adding the latter inequality across all elements $a \in \calA$ completes the proof of ineq.~\eqref{ineq5:aux_6}.
\hfill $\blacksquare$

\begin{mylemma}\label{lem:curvature2}
Consider a finite ground set $\mathcal{V}$ and a monotone set function $f:2^\mathcal{V}\mapsto \mathbb{R}$ such that $f$ is non-negative and $f(\emptyset)=0$. Then, for any sets $\mathcal{A}\subseteq \mathcal{V}$ and $\mathcal{B}\subseteq \mathcal{V}$ such that $\calA \cap \calB=\emptyset$, it holds:
\begin{equation*}
f(\mathcal{A}\cup \mathcal{B})\geq (1-c_f)\left(f(\mathcal{A})+f(\mathcal{B})\right).
\end{equation*}
\end{mylemma}
\paragraph*{Proof of Lemma~\ref{lem:curvature2}}
Let $\mathcal{B}=\{b_1, b_2, \ldots, b_{|\mathcal{B}|}\}$. Then, 
\begin{equation}
f(\mathcal{A}\cup \mathcal{B})=f(\mathcal{A})+\sum_{i=1}^{|\mathcal{B}|}f(b_i|\mathcal{A}\cup \{b_1, b_2, \ldots, b_{i-1}\}). \label{eq1:lemma_curvature2}
\end{equation} 
The definition of total curvature in Definition~\ref{def:total_curvature} implies:
\begin{align}
&\!\!\!f(b_i|\mathcal{A}\cup \{b_1, b_2, \ldots, b_{i-1}\})\geq\nonumber\\
& (1-c_f)f(b_i|\{b_1, b_2, \ldots, b_{i-1}\}). \label{eq2:lemma_curvature2}
\end{align} 
The proof is completed by substituting ineq.~\eqref{eq2:lemma_curvature2} in eq.~\eqref{eq1:lemma_curvature2} and then by taking into account that it holds $f(\mathcal{A})\geq (1-c_f)f(\mathcal{A})$, since $0\leq c_f\leq 1$.
\hfill $\blacksquare$

\begin{mylemma}\label{lem:curvature}
Consider a finite ground set $\mathcal{V}$ and a non-decreasing set function $f:2^\mathcal{V}\mapsto \mathbb{R}$ such that $f$ is non-negative and $f(\emptyset)=0$. Then, for any set $\mathcal{A}\subseteq \mathcal{V}$ and any set $\mathcal{B}\subseteq \mathcal{V}$ such that $\calA \cap \calB=\emptyset$, it holds:
\begin{equation*}
f(\mathcal{A}\cup \mathcal{B})\geq (1-c_f)\left(f(\mathcal{A})+\sum_{b \in \mathcal{B}}f(b)\right).
\end{equation*}
\end{mylemma}
\paragraph*{Proof of Lemma~\ref{lem:curvature}}
Let $\mathcal{B}=\{b_1, b_2, \ldots, b_{|\mathcal{B}|}\}$. Then, 
\begin{equation}
f(\mathcal{A}\cup \mathcal{B})=f(\mathcal{A})+\sum_{i=1}^{|\mathcal{B}|}f(b_i|\mathcal{A}\cup \{b_1, b_2, \ldots, b_{i-1}\}). \label{eq1:lemma_curvature}
\end{equation} 
In addition, Definition~\ref{def:total_curvature} of total curvature implies:
\begin{align}
f(b_i|\mathcal{A}\cup \{b_1, b_2, \ldots, b_{i-1}\})&\geq (1-c_f)f(b_i|\emptyset)\nonumber\\
&=(1-c_f)f(b_i), \label{eq2:lemma_curvature}
\end{align} 
where the latter equation holds since $f(\emptyset)=0$.
The proof is completed by substituting~\eqref{eq2:lemma_curvature} in~\eqref{eq1:lemma_curvature} and then taking into account that $f(\mathcal{A})\geq (1-c_f)f(\mathcal{A})$ since $0\leq c_f\leq 1$. \hfill $\blacksquare$

\begin{mylemma}\label{lem:subratio}
Consider a finite ground set $\mathcal{V}$ and a non-decreasing set function $f:2^\mathcal{V}\mapsto \mathbb{R}$ such that $f$ is non-negative and $f(\emptyset)=0$. Then for any set $\mathcal{A}\subseteq \mathcal{V}$ and any set  $\mathcal{B}\subseteq \mathcal{V}$ such that $\mathcal{A}\setminus\mathcal{B}\neq \emptyset$, it holds:
\begin{equation*}
f(\mathcal{A})+(1-c_f) f(\mathcal{B})\geq (1-c_f) f(\mathcal{A}\cup \mathcal{B})+f(\mathcal{A}\cap \mathcal{B}).
\end{equation*}
\end{mylemma}
\paragraph*{Proof of Lemma~\ref{lem:subratio}}
Let $\mathcal{A}\setminus\mathcal{B}=\{i_1,i_2,\ldots, i_r\}$, where $r=|\mathcal{A}-\mathcal{B}|$. From Definition~\ref{def:total_curvature} of total curvature $c_f$, for any $i=1,2, \ldots, r$\!, it is  $f(i_j|\mathcal{A} \cap \mathcal{B} \cup \{i_1, i_2, \ldots, i_{j-1}\})\geq (1-c_f) f(i_j|\mathcal{B} \cup \{i_1, i_2, \ldots, i_{j-1}\})$. Summing these $r$ inequalities,
$$f(\mathcal{A})-f(\mathcal{A}\cap \mathcal{B})\geq (1-c_f) \left(f(\mathcal{A}\cup \mathcal{B})-f(\mathcal{B})\right),$$
which implies the lemma. \hfill $\blacksquare$

\begin{mycorollary}\label{cor:ineq_from_lemmata}
Consider a finite ground set $\mathcal{V}$ and a non-decreasing set function $f:2^\mathcal{V}\mapsto \mathbb{R}$ such that $f$ is non-negative and $f(\emptyset)=0$. Then, for any set $\mathcal{A}\subseteq \mathcal{V}$ and any set $\mathcal{B}\subseteq \mathcal{V}$ such that $\mathcal{A}\cap\mathcal{B}=\emptyset$, it holds:
\begin{equation*}
f(\mathcal{A})+\sum_{b \in \mathcal{B}}f(b) \geq (1-c_f)  f(\mathcal{A}\cup \mathcal{B}).
\end{equation*}
\end{mycorollary}
\paragraph*{Proof of Corollary~\ref{cor:ineq_from_lemmata}}
 Let $\mathcal{B}=\{b_1,b_2,\ldots,b_{|\mathcal{B}|}\}$. 
\begin{align}
f(\mathcal{A})+\sum_{i=1}^{|\mathcal{B}|}f(b_i) &\geq (1-c_f) f(\mathcal{A})+\sum_{i=1}^{|\mathcal{B}|}f(b_i))\label{ineq:cor_aux1} \\
& \geq (1-c_f) f(\mathcal{A}\cup \{b_1\})+\sum_{i=2}^{|\mathcal{B}|}f(b_i)\nonumber\\
& \geq (1-c_f) f(\mathcal{A}\cup \{b_1,b_2\})+\sum_{i=3}^{|\mathcal{B}|}f(b_i)\nonumber\\
& \;\;\vdots \nonumber\\
& \geq (1-c_f) f(\mathcal{A}\cup \mathcal{B}),\nonumber
\end{align}
where~\eqref{ineq:cor_aux1} holds since $0\leq c_f\leq 1$, and the rest due to Lemma~\ref{lem:subratio} since $\mathcal{A}\cap\mathcal{B}=\emptyset$ implies $\mathcal{A}\setminus \{b_1\}\neq \emptyset$, $\mathcal{A}\cup \{b_1\}\setminus \{b_2\}\neq \emptyset$, $\ldots$, $\mathcal{A}\cup \{b_1,b_2,\ldots, b_{|\mathcal{B}|-1}\}\setminus \{b_{|\mathcal{B}|}\}\neq \emptyset$. 

\hfill $\blacksquare$

\begin{mylemma}\label{lem:adapt_res_greedy_vs_any}
Recall the notation in Algorithm~\ref{alg:rob_sub_max}. 
Given the sets $\calS_{1,1},\ldots,\calS_{T,1}$ selected by Algorithm~\ref{alg:rob_sub_max} (lines~\ref{line:sort}-\ref{line:bait} of Algorithm~\ref{alg:rob_sub_max}), then, for each $t=1,\ldots,T$\!, let the set $\calO_t$ be a subset  ---any subset--- of $\calV_t\setminus \calS_{t,1}$ of cardinality $\alpha_t-\beta_t$. Then, for the sets $\calS_{1,2},\ldots,\calS_{T,2}$ selected by Algorithm~\ref{alg:rob_sub_max} (lines~\ref{line:begin_while}-\ref{line:end_while} of Algorithm~\ref{alg:rob_sub_max}), it holds:  
\begin{equation}\label{eq:adapt_res_greedy_vs_any}
f(\calS_{1,2},\ldots,\calS_{T,2})\geq (1-c_f)^2f(\calO_1,\ldots,\calO_T).
\end{equation}
\end{mylemma}
\paragraph*{Proof of Lemma~\ref{lem:adapt_res_greedy_vs_any}}
For all $t=1,2,\ldots,T$\!, let the set $\calR_t\triangleq \calA_t\setminus \calB_t$; namely, $\calR_t$ is the set that remains after the optimal (worst-case) removal $\calB_t$ from $\calA_t$.  Furthermore, let the element $s_{t,2}^i\in \calS_{t,2}$ denote the $i$-th element added in $\calS_{t,2}$ per the greedy subroutine in lines~\ref{line:begin_while}-\ref{line:end_while} of Algorithm~\ref{alg:rob_sub_max}; i.e.,  $\calS_{t,2}=\{s_{t,2}^1,\ldots,s_{t,2}^{\alpha_t-\beta_t}\}$.  In addition,  for all $i=1,\ldots, \alpha_t-\beta_t$, denote $\calS_{t,2}^i\triangleq\{s_{t,2}^1,\ldots,s_{t,2}^{i}\}$, and also set $\calS_{t,2}^{0}\triangleq \emptyset$.  Next, order the elements in each $\calO_t$ so that $\calO=\{o_t^1,\ldots,o_t^{\alpha_t-\beta_t}\}$ and so that if $o_t^i$ is also in $\calS_{t,2}$, then it holds $o_t^i=s_{t,2}^i$; i.e., order the elements in each $\calO_t$ so that the common elements in $\calO_t$ and $\calS_{t,2}$ appear at the same index.
Moreover, for all $i=1,\ldots, \alpha_t-\beta_t$, denote $\calO_{t}^i\triangleq\{o_{t}^1,\ldots,o_{t}^{i}\}$, and also set $\calO_{t}^{0}\triangleq \emptyset$.  Finally, let: $\calO_{1:t}\triangleq \calO_1\cup\ldots\cup \calO_{t}$; $\calO_{1:0}\triangleq\emptyset$; $\calS_{1:t,2}\triangleq \calS_{1,2}\cup\ldots\cup \calS_{t,2}$; and $\calS_{1:0,2}\triangleq\emptyset$. Then, it holds:
\begin{align}
&\!\!\!f(\calO_1,\ldots,\calO_T)\nonumber\\
&=\sum_{t=1}^{T}\sum_{i=1}^{\alpha_t-\beta_t}f(o_t^i|\calO_{1:t-1}\cup \calO_t^{i-1})\label{aux10:1}\\
&\leq \frac{1}{1-c_f}\sum_{t=1}^{T}\sum_{i=1}^{\alpha_t-\beta_t}f(o_t^i|\calR_{1:t-1}\cup \calS_{t,2}^{i-1})\label{aux10:2}\\
&\leq \frac{1}{1-c_f}\sum_{t=1}^{T}\sum_{i=1}^{\alpha_t-\beta_t}f(s_{t,2}^i|\calR_{1:t-1}\cup \calS_{t,2}^{i-1})\label{aux10:3}\\
&\leq \frac{1}{(1-c_f)^2}\sum_{t=1}^{T}\sum_{i=1}^{\alpha_t-\beta_t}f(s_{t,2}^i|\calS_{1:t-1,2}\cup \calS_{t,2}^{i-1})\label{aux10:4}\\
%
&=\frac{1}{(1-c_f)^2}f(\calS_{1,2},\ldots,\calS_{T,2})\label{aux10:5}.
\end{align}
where the eqs.~\eqref{aux10:1}-\eqref{aux10:5} hold for the following reasons: eq.~\eqref{aux10:1} holds due the notation introduced in eq.~\eqref{notation:marginal}; ineq.~\eqref{aux10:2} holds since Definition~\ref{def:total_curvature} of total curvature implies ineq.~\eqref{eq:ineq_total_curvature}, and since the definition of each $o_t^i$ implies that because $o_t^i \notin \calO_t^{i-1}$\!\!, then it also is $o_t^i \notin \calS_{t,2}^{i-1}$\!\!, and as a result, because $o_t^i \notin \calO_{1:t-1}\cup\calO_t^{i-1}$\!\!, then it also is $o_t^i \notin \calR_{1:t-1}\cup\calS_{t,2}^{i-1}$ (which fact allows the application of ineq.~\eqref{eq:ineq_total_curvature}); ineq.~\eqref{aux10:3} holds since the element $s_{t,2}^i$ is chosen greedily, given $\calR_{1:t-1}\cup \calS_{t,2}^{i-1}$; ineq.~\eqref{aux10:4} holds for the same reasons as ineq.~\eqref{aux10:2}; similarly, eq.~\eqref{aux10:5} holds for the same reasons as eq.~\eqref{aux10:1}.
\hfill $\blacksquare$

\begin{algorithm}[t]
\caption{Local greedy algorithm~\cite[Section~4]{fisher1978analysis}.}
\begin{algorithmic}[1]
\REQUIRE 
Integer $T$; finite ground sets $\mathcal{K}_1,\ldots,\calK_T$; set function $f:2^{\mathcal{K}_1}\times \cdots\times 2^{\mathcal{K}_T} \mapsto \mathbb{R}$ such that $f$ is non-decreasing, non-negative, and $f(\emptyset)=0$; integers $\delta_1,\ldots,\delta_T$ such that $0\leq\delta_t \leq |\mathcal{K}_t|$, for all $t=1,\ldots,T$\!.
\ENSURE  At each step $t=1,2,\ldots,T$\!, set $\mathcal{M}_{t}$.
\medskip

\FORALL {$t=1,\ldots,T$}
\STATE $\mathcal{M}_{t}\leftarrow\emptyset$;
\WHILE {$|\mathcal{M}_{t}| < \delta_t$} \label{line2:begin_while} 
\STATE $x\in \arg\max_{y \in \mathcal{K}_t\setminus \mathcal{M}_{t}}f(\calS_{1},\ldots,\calS_{t-1}, \mathcal{M}_{t}\cup \{y\})$; \label{line2:greedy_1}
\STATE $\mathcal{M}_{t} \leftarrow \{x\}\cup \mathcal{M}_{t}$;\label{line2:greedy_2}
\ENDWHILE \label{line2:end_while}
\ENDFOR 

\end{algorithmic}\label{alg:local}
\end{algorithm}

\begin{mylemma}\label{lem:local_greedy_vs_opt}
Recall the notation in Algorithm~\ref{alg:rob_sub_max}; in particular, consider the sets $\calS_{1,1},\ldots,\calS_{T,1}$ selected by Algorithm~\ref{alg:rob_sub_max} (lines~\ref{line:sort}-\ref{line:bait} of Algorithm~\ref{alg:rob_sub_max}).  Moreover, consider the notation in Algorithm~\ref{alg:local},\footnote{The local greedy Algorithm~\ref{alg:local} is connected to Algorithm~\ref{alg:rob_sub_max} as follows: Algorithm~\ref{alg:rob_sub_max} reduces to Algorithm~\ref{alg:local} if in Problem~\ref{pr:robust_sub_max} we assume no removals; equivalently, if in Algorithm~\ref{alg:rob_sub_max} we assume that for all $t=1,\ldots,T$ it is $\calB_t=\emptyset$ (no attacks), and correspondingly, that $\beta_t=0$, which implies $\calS_{t,1}=\emptyset$.
} and
for all $t=1,2,\ldots,T$\!, let in Algorithm~\ref{alg:local} be $\calK_t=\calV_t\setminus \calS_{t,1}$ and $\delta_t=\alpha_t-\beta_t$. 
Finally, for all $t=1,2,\ldots,T$\!, let the set $\calP_t$ be such that $\calP_t\subseteq \calK_t$, $|\calP_t|\leq\delta_t$, and $f({\calP}_1,\ldots,{\calP}_T)$ is maximal, that is: 
\begin{align}\label{eq:local_greedy}
\begin{split}
&(\calP_1,\ldots,\calP_T)\in \\
&\arg \max_{\bar{\calP}_1\subseteq \calK_1, |\bar{\calP}_1|\leq \delta_1}\cdots \max_{\bar{\calP}_T\subseteq \calK_T, |\bar{\calP}_T|\leq \delta_T} f(\bar{\calP}_1,\ldots,\bar{\calP}_T).
\end{split}
\end{align} Then, it holds:
\begin{equation}\label{eq:local_greedy_vs_opt}
f(\calM_1,\calM_2,\ldots,\calM_T)\geq (1-c_f)f(\calP_1,\calP_2,\ldots,\calP_T).
\end{equation}
\end{mylemma}
\paragraph*{Proof of Lemma~\ref{lem:local_greedy_vs_opt}}
We use similar notation to the one introduced in the proof of Lemma~\ref{lem:adapt_res_greedy_vs_any}.  In addition, ---again similarly to the proof of Lemma~\ref{lem:adapt_res_greedy_vs_any},--- we order the elements in each $\calP_t$ so that $\calP_t=\{p_t^1,\ldots,p_T^{\delta_t}\}$ and so that they appear in the same place as in $\calM_t$. Moreover, we let the element $m_t^i\in\calM_t$ denote the $i$-th element added in $\calM_t$ per the greedy subroutine in lines~\ref{line2:begin_while}-\ref{line2:end_while} of Algorithm~\ref{alg:local}.  Then, it holds:
\begin{align}
&\!\!\!f(\calP_1,\calP_2,\ldots,\calP_T)\nonumber\\
&=\sum_{t=1}^{T}\sum_{i=1}^{\delta_t}f(p_t^i|\calP_{1:t-1}\cup \calP_t^{i-1})\label{aux30:1}\\
&\leq \frac{1}{1-c_f}\sum_{t=1}^{T}\sum_{i=1}^{\delta_t}f(p_t^i|\calM_{1:t-1}\cup \calM_t^{i-1})\label{aux30:2}\\
&\leq \frac{1}{1-c_f}\sum_{t=1}^{T}\sum_{i=1}^{\delta_t}f(m_t^i|\calM_{1:t-1}\cup \calM_t^{i-1})\label{aux30:3}\\
&=\frac{1}{1-c_f}f(\calM_1,\calM_2,\ldots,\calM_T).\label{aux30:4}
\end{align}
where the eqs.~\eqref{aux30:1}-\eqref{aux30:4} hold for the following reasons: eq.~\eqref{aux30:1} holds due to the notation introduced in eq.~\eqref{notation:marginal}; ineq.~\eqref{aux30:2} holds since Definition~\ref{def:total_curvature} of total curvature implies ineq.~\eqref{eq:ineq_total_curvature}, and since the definition of each $p_t^i$ implies that because $p_t^i \notin \calP_t^{i-1}$\!\!, then it also is $p_t^i \notin \calM_{t}^{i-1}$\!\!, and as a result, because $p_t^i \notin \calP_{1:t-1}\cup\calP_t^{i-1}$\!\!, then it also is $p_t^i \notin \calM_{1:t-1}\cup\calM_{t}^{i-1}$ (which fact allows the application of ineq.~\eqref{eq:ineq_total_curvature}); ineq.~\eqref{aux30:3} holds since the element $m_{t}^i$ is chosen greedily, given $\calM_{1:t-1}\cup \calM_{t}^{i-1}$; eq.~\eqref{aux30:4} holds for the same reasons as eq.~\eqref{aux30:1}.
\hfill $\blacksquare$

\begin{mycorollary}\label{lem:adapt_res_greedy_vs_opt_local_greedy}
Recall the notation in Algorithm~\ref{alg:rob_sub_max}.
In particular, consider the sets $\calS_{1,1},\ldots,\calS_{T,1}$ selected by Algorithm~\ref{alg:rob_sub_max} (lines~\ref{line:sort}-\ref{line:bait} of Algorithm~\ref{alg:rob_sub_max}), as well as, the sets $\calS_{1,2},\ldots,\calS_{T,2}$ selected by Algorithm~\ref{alg:rob_sub_max} (lines~\ref{line:begin_while}-\ref{line:end_while} of Algorithm~\ref{alg:rob_sub_max}).  Finally, per the notation of Lemma~\ref{lem:local_greedy_vs_opt}, for all $t=1,2,\ldots,T$\!, consider $\calK_t=\calV_t\setminus\calS_{t,1}$ and $\delta_t=\alpha_t-\beta_t$, and let the set $\calP_t$ be such that $\calP_t\subseteq \calK_t$, $|\calP_t|\leq\delta_t$, and $f({\calP}_1,\ldots,{\calP}_T)$ is maximal, per eq.~\eqref{eq:local_greedy}. Then, for the sets $\calS_{1,2},\ldots,\calS_{T,2}$ selected by Algorithm~\ref{alg:rob_sub_max} (lines~\ref{line:begin_while}-\ref{line:end_while} of Algorithm~\ref{alg:rob_sub_max}), it holds:  
\begin{equation}\label{ineq:main_adapt_res_greedy_vs_opt_local_greedy}
f(\calS_{1,2},\ldots,\calS_{T,2})\geq (1-c_f)^3f(\calP_1,\calP_2,\ldots,\calP_T).
\end{equation}
\end{mycorollary}
\paragraph*{Proof of Corollary~\ref{lem:adapt_res_greedy_vs_opt_local_greedy}}
The proof follows from Lemma~\ref{lem:adapt_res_greedy_vs_any} and Lemma~\ref{lem:local_greedy_vs_opt}. In particular, let $\calO_t=\calM_t$ in ineq.~\eqref{eq:adapt_res_greedy_vs_any} to get:
\begin{equation}\label{eq:adapt_res_greedy_vs_opt_local_greedy}
f(\calS_{1,2},\ldots,\calS_{T,2})\geq (1-c_f)^2f(\calM_1,\ldots,\calM_T).
\end{equation}
Using in ineq.~\eqref{eq:adapt_res_greedy_vs_opt_local_greedy} the ineq.~\eqref{eq:local_greedy_vs_opt}, the proof is complete.
\hfill $\blacksquare$

\begin{mylemma}\label{lem:from_max_to_minmax}
Recall the notation in Theorem~\ref{th:alg_rob_sub_max_performance}. In addition, per the notation of Corollary~\ref{lem:adapt_res_greedy_vs_opt_local_greedy}, for all $t=1,2,\ldots,T$\!, consider $\calK_t=\calV_t\setminus\calS_{t,1}$ and $\delta_t=\alpha_t-\beta_t$, and let the set $\calP_t$ be such that $\calP_t\subseteq \calK_t$, $|\calP_t|\leq\delta_t$, and $f({\calP}_1,\ldots,{\calP}_T)$ is maximal, per eq.~\eqref{eq:local_greedy}. Then, it holds:
\begin{equation}\label{eq:toprovefrom_max_to_minmax}
f(\calP_1,\ldots,\calP_T)\geq f(\calA_{1:T}^\star\setminus \calB^\star(\calA_{1:T}^\star)).
\end{equation}
\end{mylemma}
\paragraph*{Proof of Lemma~\ref{lem:from_max_to_minmax}}
Consider the following notation: since for each  $t=1,\ldots,T$\!, it is $\calK_t=\calV_t\setminus\calS_{t,1}$, let:
\begin{align}
\begin{split}
&\!\!\!h(\calS_{1,1},\ldots,\calS_{T,1})\triangleq\\
&\max_{\bar{\calP}_1\subseteq \calV_1\setminus\calS_{1,1}, |\bar{\calP}_1|\leq \delta_1}\cdots \max_{\bar{\calP}_T\subseteq \calV_T\setminus\calS_{T,1}, |\bar{\calP}_T|\leq \delta_T} f(\bar{\calP}_1,\ldots,\bar{\calP}_T).
\end{split}
\end{align}

Given the above notation, for any $\hat{\calP}_1,\ldots,\hat{\calP}_T$ such that for all $t=1,\ldots,T$ it is $\hat{\calP}_t\subseteq \calV_t\setminus\calS_{t,1}$ and $|\hat{\calP}_t|\leq \delta_t$, it holds:
\begin{align}
&h(\calS_{1,1},\ldots,\calS_{T,1})\geq f(\hat{\calP}_1,\ldots,\hat{\calP}_T)\Rightarrow\label{aux40:0}\\
&h(\calS_{1,1},\ldots,\calS_{T,1})\geq\nonumber\\
&\hspace{2cm}\max_{\bar{\calP}_T\subseteq \calV_T\setminus\calS_{T,1}, |\bar{\calP}_T|\leq \delta_T}f(\hat{\calP}_1,\ldots,\hat{\calP}_{T-1},\bar{\calP}_T)\Rightarrow\nonumber\\
&\min_{\bar{\calB}_{T}\subseteq \calV_T, |\bar{\calB}_{T}|\leq \beta_T}h(\calS_{1,1},\ldots,\calS_{T-1,1},\bar{\calB}_{T})\geq\nonumber\\
&\hspace{0cm}\min_{\bar{\calB}_{T}\subseteq \calV_T, |\bar{\calB}_{T}|\leq \beta_T}\max_{\bar{\calP}_T\subseteq \calV_T\setminus\bar{\calB}_{T}, |\bar{\calP}_T|\leq \delta_T}f(\hat{\calP}_1,\ldots,\hat{\calP}_{T-1},\bar{\calP}_T).\label{aux40:1}
\end{align}
Denote the right-hand-side of ineq.~\eqref{aux40:1} by $z(\hat{\calP}_1,\ldots,\hat{\calP}_{T-1})$. Since  $\delta_T=\alpha_T-\beta_T$, and for $\bar{\calP}_T$ in  ineq.~\eqref{aux40:1} it is $\bar{\calP}_T\subseteq \calV_T\setminus\bar{\calB}_{T}$ and $|\bar{\calP}_T|\leq \delta_T$, then it equivalently holds:
\begin{align}
&z(\hat{\calP}_1,\ldots,\hat{\calP}_{T-1})=\nonumber\\
&\min_{\bar{\calB}_{T}\subseteq \calV_T, |\bar{\calB}_{T}|\leq \beta_T}\max_{\bar{\calA}_T\subseteq \calV_T, |\bar{\calA}_T|\leq \alpha_T}f(\hat{\calP}_1,\ldots,\hat{\calP}_{T-1},\bar{\calA}_T\setminus \bar{\calB}_{T}).\label{aux40:2}
\end{align}
Let in ineq.~\eqref{aux40:2} be $w(\bar{\calA}_T\setminus \bar{\calB}_{T})\triangleq f(\hat{\calP}_1,\ldots,\hat{\calP}_{T-1},\bar{\calA}_T\setminus \bar{\calB}_{T})$. We prove next that it holds:
\begin{align}
&z(\hat{\calP}_1,\ldots,\hat{\calP}_{T-1})\geq \nonumber \\
&\max_{\bar{\calA}_T\subseteq \calV_T, |\bar{\calA}_T|\leq \alpha_T}\min_{\bar{\calB}_{T}\subseteq \calV_T, |\bar{\calB}_{T}|\leq \beta_T}w(\bar{\calA}_T\setminus \bar{\calB}_{T}).\label{aux40:3}
\end{align}
The proof of ineq.~\eqref{aux40:3} is as follows: for any $\hat{\calA}_T\subseteq \calV_T, |\hat{\calA}_T|\leq \alpha_T$, and any $\hat{\calS}_{T,1}\subseteq \calV_T, |\hat{\calS}_{T,1}|\leq \beta_T$, it holds:
\begin{align}
\max_{\bar{\calA}_T\subseteq \calV_T, |\bar{\calA}_T|\leq \alpha_T} w(\bar{\calA}_T\setminus \hat{\calS}_{T,1})\geq w(\hat{\calA}_T\setminus \hat{\calS}_{T,1}) &\Rightarrow \nonumber\\
\min_{\bar{\calB}_{T}\subseteq \calV_T, |\bar{\calB}_{T}|\leq \beta_T}\max_{\bar{\calA}_T\subseteq \calV_T, |\bar{\calA}_T|\leq \alpha_T} w(\bar{\calA}_T\setminus \bar{\calB}_{T})&\geq \nonumber\\
&\hspace{-3cm}\min_{\bar{\calB}_{T}\subseteq \calV_T, |\bar{\calB}_{T}|\leq \beta_T}w(\hat{\calA}_T\setminus \bar{\calB}_{T}),\label{aux40:4}
\end{align}
and now ineq.~\eqref{aux40:4} implies ineq.~\eqref{aux40:2}.  
Overall, ineq.~\eqref{aux40:1} becomes:
\begin{align}
&\min_{\bar{\calB}_{T}\subseteq \calV_T, |\bar{\calB}_{T}|\leq \beta_T}h(\calS_{1,1},\ldots,\calS_{T-1,1},\bar{\calB}_{T})\geq\nonumber\\
&\max_{\bar{\calA}_T\subseteq \calV_T, |\bar{\calA}_T|\leq \alpha_T}\min_{\bar{\calB}_{T}\subseteq \calV_T, |\bar{\calB}_{T}|\leq \beta_T} f(\hat{\calP}_1,\ldots,\hat{\calP}_{T-1},\bar{\calA}_T\setminus \bar{\calB}_{T}).\label{aux40:5}
\end{align}
The left-hand-side of ineq.~\eqref{aux40:5} is a function of $\calS_{1,1},\ldots,\calS_{T-1,1}$; denote it as $h'(\calS_{1,1},\ldots,\calS_{T-1,1})$.  Similarly, the right-hand-side of ineq.~\eqref{aux40:5} is a function of $\hat{\calP}_1,\ldots,\hat{\calP}_{T-1}$; denote it as $f'(\hat{\calP}_1,\ldots,\hat{\calP}_{T-1})$. Given these notations, ineq.~\eqref{aux40:5} is equivalently written as:
\begin{align}
&h'(\calS_{1,1},\ldots,\calS_{T-1,1})\geq f'(\hat{\calP}_1,\ldots,\hat{\calP}_{T-1}),\label{aux40:6}
\end{align}
which has the same form as ineq.~\eqref{aux40:0}.  Therefore, following the same steps as those we used starting from ineq.~\eqref{aux40:0} to prove ineq.~\eqref{aux40:5}, it holds:
\begin{align}
&\min_{\bar{\calB}_{T-1}\subseteq \calV_{T-1}, |\bar{\calB}_{T-1}|\leq \beta_{T-1}}h'(\calS_{1,1},\ldots,\calS_{T-2,1},\bar{\calB}_{T-1})\geq\nonumber\\
&\max_{\bar{\calA}_{T-1}\subseteq \calV_{T-1}, |\bar{\calA}_{T-1}|\leq \alpha_{T-1}}\min_{\bar{\calB}_{T-1}\subseteq \calV_{T-1}, |\bar{\calB}_{T-1}|\leq \beta_{T-1}}\nonumber\\ &\hspace{3cm}f'(\hat{\calP}_1,\ldots,\hat{\calP}_{T-2},
\bar{\calA}_{T-1}\setminus \bar{\calB}_{T-1}),\label{aux40:7}
\end{align}
which has the same form as ineq.~\eqref{aux40:5}.  Repeating the same steps  as those we used starting from ineq.~\eqref{aux40:0} to prove ineq.~\eqref{aux40:5} for another $T-2$ times, it holds:
\begin{align}
&\min_{\bar{\calB}_{1,}\subseteq \calV_1, |\bar{\calB}_{1}|\leq \beta_1}\cdots\min_{\bar{\calB}_{T}\subseteq \calV_T, |\bar{\calB}_{T}|\leq \beta_T}h(\bar{\calB}_{1},\ldots\bar{\calB}_{T})\geq\nonumber\\
&\max_{\bar{\calA}_{1}\subseteq \calV_{1}, |\bar{\calA}_{1}|\leq \alpha_{1}}\min_{\bar{\calB}_{1}\subseteq \calV_{1}, |\bar{\calB}_{1}|\leq \beta_{1}}\!\!\cdots\!\!\max_{\bar{\calA}_{T}\subseteq \calV_{T}, |\bar{\calA}_{T}|\leq \alpha_{T}}\min_{\bar{\calB}_{1}\subseteq \calV_{T}, |\bar{\calB}_{T}|\leq \beta_{T}}\nonumber\\ &\hspace{5cm}f(\bar{\calA}_{1}\setminus \bar{\calB}_{1},\ldots,
\bar{\calA}_{T}\setminus \bar{\calB}_{T}),\label{aux40:40}
\end{align}
which is implies ineq.~\eqref{eq:toprovefrom_max_to_minmax}, since the right-hand-side of ineq.~\eqref{aux40:40} is equal to the right-hand-side of ineq.~\eqref{eq:toprovefrom_max_to_minmax}, and  ---with respect now to the left-hand-side of ineq.~\eqref{aux40:40}--- it is:
\belowdisplayskip=-9pt\begin{align*}
&\!\!\!f(\calP_1,\ldots,\calP_T)\geq\\
&\min_{\bar{\calB}_{1,}\subseteq \calV_1, |\bar{\calB}_{1}|\leq \beta_1}\cdots\min_{\bar{\calB}_{T}\subseteq \calV_T, |\bar{\calB}_{T}|\leq \beta_T}h(\bar{\calB}_{1},\ldots\bar{\calB}_{T}).
\end{align*}
\hfill $\blacksquare$

\section{Proof of Theorem~\ref{th:alg_rob_sub_max_performance}}

We first prove Theorem~\ref{th:alg_rob_sub_max_performance}'s part 1 (approximation performance), and then, Theorem~\ref{th:alg_rob_sub_max_performance}'s part 2 (running time).

\subsection{Proof of Theorem~\ref{th:alg_rob_sub_max_performance}'s part 1 (approximation performance)}

\begin{figure}[t]
\def \setAone{ (0,0) circle (1cm) }
\def \setBone{ (.5,0) circle (0.4cm)}
\def \setAtwo{ (2.5,0) circle (1cm) }
\def \setBtwo{ (3.0,0) circle (0.4cm)}
\def \myrectangle{ (-1.5, -1.5) rectangle (4, 1.5) }
\begin{center}
\begin{tikzpicture}
\draw \myrectangle node[below left]{$\mathcal{V}_t$};
\draw \setAone node[left]{$\mathcal{S}_{t,1}$};
\draw \setBone node[]{$\mathcal{B}_{t,1}^\star$};
\draw \setAtwo node[left]{$\mathcal{S}_{t,2}$};
\draw \setBtwo node[]{$\mathcal{B}_{t,2}^\star$};
\end{tikzpicture}
\end{center}
\caption{Venn diagram, where the sets $\mathcal{S}_{t,1}, \mathcal{S}_{t,2},\mathcal{B}_{t,1}^\star, \mathcal{B}_{t,2}^\star$ are as follows: per Algorithm~\ref{alg:rob_sub_max}, $\mathcal{S}_{t,1}$  and $\mathcal{S}_{t,2}$ are such that $\mathcal{A}_t=\mathcal{S}_{t,1}\cup \mathcal{S}_{t,2}$.  In addition, due to their construction, it holds  $\mathcal{S}_{t,1}\cap \mathcal{S}_{t,2}=\emptyset$. Next, 
$\mathcal{B}_{t,1}^\star$ and $\mathcal{B}_{t,2}^\star$ are such that  $\mathcal{B}_{t,1}^\star=\mathcal{B}^\star(\mathcal{A}_{1:T})\cap\mathcal{S}_{t,1}$, and $\mathcal{B}_2^\star=\mathcal{B}^\star(\mathcal{A}_{1:T})\cap\mathcal{S}_{t,2}$; therefore, it is $\mathcal{B}_{t,1}^\star\cap \mathcal{B}_{t,2}^\star=\emptyset$ and $\mathcal{B}^\star(\mathcal{A}_{1:T})=(\mathcal{B}_{1,1}^\star\cup \mathcal{B}_{1,2}^\star)\cup \cdots\cup(\mathcal{B}_{T,1}^\star\cup \mathcal{B}_{T,2}^\star)$.  
}\label{fig:venn_diagram_for_proof}
\end{figure}
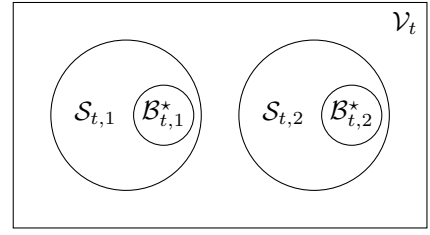

We first prove ineq.~\eqref{ineq:bound_non_sub}; then, we prove ineq.~\eqref{ineq:bound_sub}.

To the above ends, we use the following notation (along with the notation introduced in Algorithm~\ref{alg:rob_sub_max}, Theorem~\ref{th:alg_rob_sub_max_performance}, and in Appendix~\ref{app:notation}): for each $t=1,\ldots,T$:
\begin{itemize}
\item let $\calS_{t,1}^+\triangleq \calS_{t,1}\setminus \calB^\star(\calA_{1:T})$, i.e., $\calS_{t,1}^+$ is the set of remaining elements in the set $\calS_{t,1}$ after the removal from $\calS_{t,1}$ of the elements in the optimal (worst-case) removal $\calB^\star(\calA_{1:T})$;
\item let $\calS_{t,2}^+\triangleq \calS_{t,2}\setminus \calB^\star(\calA_{1:T})$, i.e., $\calS_{t,2}^+$ is the set of remaining elements in the set $\calS_{t,2}$ after the removal from $\calS_{t,2}$ of the elements in the optimal (worst-case) removal $\calB^\star(\calA_{1:T})$;
\item let the sets $\calP_1,\ldots,\calP_T$ be a solution to the maximization problem in eq.~\eqref{eq:local_greedy} per the conditions in Corollary~\ref{lem:adapt_res_greedy_vs_opt_local_greedy}, i.e., for $\calK_t=\calV_t\setminus \calS_{t,1}$ and $\delta_t=\alpha_t-\beta_t$. 
\end{itemize}

\paragraph*{Proof of ineq.~\eqref{ineq:bound_non_sub}}  Consider that the objective function~$f$ is non-decreasing and such that (without loss of generality) $f$ is non-negative and $f(\emptyset)=0$. Then, the proof of ineq.~\eqref{ineq:bound_non_sub} follows by making the following observations:
\belowdisplayskip=8pt\begin{align}
&\!\!\!f(\calA_{1:T}\setminus \calB^\star(\calA_{1:T}))\nonumber\\
&=f(\calS_{1,1}^+\cup \calS_{1,2}^+,\ldots,\calS_{T,1}^+\cup \calS_{T,2}^+)\label{ineq2:aux_14}\\
&\geq (1-c_f)\sum_{t=1}^T\sum_{v\in\calS_{t,1}^+\cup \calS_{t,2}^+}f(v)\label{ineq2:aux_15}\\
&\geq (1-c_f)\sum_{t=1}^T\sum_{v\in\calS_{t,2}}f(v)\label{ineq2:aux_16}\\
&\geq (1-c_f)^2f(\calS_{1,2},\ldots,\calS_{T,2})\label{ineq2:aux_17}\\
&\geq (1-c_f)^5f(\calP_1,\ldots,\calP_T)\label{ineq2:aux_18}\\
&\geq (1-c_f)^5f(\calA_{1:T}^\star\setminus \calB^\star(\calA_{1:T}^\star)),\label{ineq2:aux_19}
\end{align}
where eqs.~\eqref{ineq2:aux_14} to~\eqref{ineq2:aux_19} hold for the following reasons: eq.~\eqref{ineq2:aux_14} follows from the definitions of the sets~$\calS_{t,1}^+$ and $\calS_{t,2}^+$; ineq.~\eqref{ineq2:aux_15} follows from ineq.~\eqref{ineq2:aux_14} due to Lemma~\ref{lem:curvature}; ineq.~\eqref{ineq2:aux_16} follows from ineq.~\eqref{ineq2:aux_15} because for all elements $v \in \calS_{t,1}^+$ and all elements  $v' \in \calS_{t,2}\setminus \calS_{t,2}^+$ it is $f(v)\geq f(v')$ (note that due to the definitions of the sets~$\calS_{t,1}^+$ and $\calS_{t,2}^+$ it is $|\calS_{t,1}^+|=|\calS_{t,2}\setminus \calS_{t,2}^+|$, that is, the number of non-removed elements in $\calS_{t,1}$ is equal to the number of removed elements in $\calS_{t,2}$), and because $\calS_{t,2}=(\calS_{t,2}\setminus \calS_{t,2}^+)\cup \calS_{t,2}^+$; ineq.~\eqref{ineq2:aux_17} follows from ineq.~\eqref{ineq2:aux_16} due to Corollary~\ref{cor:ineq_from_lemmata}; ineq.~\eqref{ineq2:aux_18} follows from ineq.~\eqref{ineq2:aux_17} due to Corollary~\ref{lem:adapt_res_greedy_vs_opt_local_greedy}; finally, ineq.~\eqref{ineq2:aux_19} follows from ineq.~\eqref{ineq2:aux_18} due to Lemma~\ref{lem:from_max_to_minmax}.  The above conclude the proof of ineq.~\eqref{ineq:bound_non_sub}.
\hfill $\blacksquare$

\paragraph*{Proof of ineq.~\eqref{ineq:bound_sub}}
Consider that the objective function~$f$ is non-decreasing submodular and such that (without loss of generality) $f$ is non-negative and $f(\emptyset)=0$.
To prove 
ineq.~\eqref{ineq:bound_sub} we follow similar observations to the ones we followed in the proof of ineq.~\eqref{ineq:bound_non_sub}; in particular:
\begin{align}
&\!\!\!f(\calA_{1:T}\setminus \calB^\star(\calA_{1:T}))\nonumber\\
&=f(\calS_{1,1}^+\cup \calS_{1,2}^+,\ldots,\calS_{T,1}^+\cup \calS_{T,2}^+)\label{ineq5:aux_14}\\
&\geq (1-\kappa_f)\sum_{t=1}^T\sum_{v\in\calS_{t,1}^+\cup \calS_{t,2}^+}f(v)\label{ineq5:aux_15}
\end{align}

\begin{align}
&\geq (1-\kappa_f)\sum_{t=1}^T\sum_{v\in\calS_{t,2}}f(v)\label{ineq5:aux_16}\\
&\geq (1-\kappa_f)f(\calS_{1,2},\ldots,\calS_{T,2})\label{ineq5:aux_17}\\
&\geq (1-\kappa_f)^4f(\calP_1,\ldots,\calP_T)\label{ineq5:aux_18}\\
&\geq (1-\kappa_f)^4 f(\calA_{1:T}^\star\setminus \calB^\star(\calA_{1:T}^\star)),\label{ineq5:aux_19}
\end{align}
where eqs.~\eqref{ineq5:aux_14} to~\eqref{ineq5:aux_19} hold for the following reasons: eq.~\eqref{ineq5:aux_14} follows from the definitions of the sets~$\calS_{t,1}^+$ and $\calS_{t,2}^+$; ineq.~\eqref{ineq5:aux_15} follows from ineq.~\eqref{ineq5:aux_14} due to Lemma~\ref{lem:non_total_curvature}; ineq.~\eqref{ineq5:aux_16} follows from ineq.~\eqref{ineq5:aux_15} because for all elements $v \in \calS_{t,1}^+$ and all elements  $v' \in \calS_{t,2}\setminus \calS_{t,2}^+$ it is $f(v)\geq f(v')$ (note that due to the definitions of the sets~$\calS_{t,1}^+$ and $\calS_{t,2}^+$ it is $|\calS_{t,1}^+|=|\calS_{t,2}\setminus \calS_{t,2}^+|$, that is, the number of non-removed elements in $\calS_{t,1}$ is equal to the number of removed elements in $\calS_{t,2}$), and because $\calS_{t,2}=(\calS_{t,2}\setminus \calS_{t,2}^+)\cup \calS_{t,2}^+$; ineq.~\eqref{ineq5:aux_17} follows from ineq.~\eqref{ineq5:aux_16} because the set function $f$ is submodular and, as~a result, the~submodularity Definition~\ref{def:sub} implies that for any sets $\mathcal{S}\subseteq \mathcal{V}$ and $\mathcal{S}'\subseteq \mathcal{V}$, it is  $f(\mathcal{S})+f(\mathcal{S}')\geq f(\mathcal{S}\cup \mathcal{S}')$~\cite[Proposition 2.1]{nemhauser78analysis};  ineq.~\eqref{ineq5:aux_18} follows from ineq.~\eqref{ineq5:aux_17} due to Corollary~\ref{lem:adapt_res_greedy_vs_opt_local_greedy}, along with the fact that since $f$ is monotone submodular it is $c_f=\kappa_f$, per Definition~\ref{def:total_curvature} of total curvature; finally, ineq.~\eqref{ineq5:aux_19} follows from ineq.~\eqref{ineq5:aux_18} due to Lemma~\ref{lem:from_max_to_minmax}.  The above conclude the proof of the $(1-\kappa_f)^4$ part in ineq.~\eqref{ineq:bound_sub}.
\hfill $\blacksquare$


\subsection{Proof of Theorem~\ref{th:alg_rob_sub_max_performance}'s part 2 (running time)}

We follow the proof of \cite[Part 2 of Theorem~\ref{th:alg_rob_sub_max_performance}]{tzoumas2017resilient}.  In~particular, we complete the proof in two steps, where we denote the time for each evaluation of the objective function $f$ as $\tau_f$: for each $t=1,\ldots,T$\!, we first compute the time line~3 of Algorithm~\ref{alg:rob_sub_max} needs to be executed, and then the time lines 5-8 of Algorithm~\ref{alg:rob_sub_max} need to be executed: \mbox{line 3} needs $|\calV_t|\tau_f+|\calV_t|\log(|\calV_t|)+|\calV_t|+O(\log(|\calV_t|))$ time, since it asks for $|\calV_t|$ evaluations of $f$\!, and their sorting, which takes $|\calV_t|\log(|\calV_t|)+|\calV_t|+O(\log(|\calV_t|))$ time, using, e.g., the merge sort algorithm.  Lines 5-8 need $(\alpha_t-\beta_t)[|\calV_t|\tau_f+|\calV_t|]$ time, 
since the while loop is repeated $\alpha_t-\beta_t$ times, and during each loop at most $|\calV_t|$ evaluations of $f$ are needed by line~5, as well as, at most $|\calV_t|$ time-steps for a maximal element in line~6
to be found.  Overall, Algorithm~\ref{alg:rob_sub_max} runs in $(\alpha_t-\beta_t)[|\calV_t|\tau_f+|\calV_t|]+|\calV_t|\log(|\calV_t|)+|\calV_t|+O(\log(|\calV_t|))=O(|\calV_t|(\alpha_t-\beta_t)\tau_f)$ time. 
\hfill $\blacksquare$

\end{document}